# A Building as a Repository: KIR, a Typed Intermediate Representation for Agent-Authored Building Information Models

**Dmitry Kuklev**
Independent Researcher · dima.kuklev9797@gmail.com

**Abstract.** Autonomous agents that author building information models need more than access to a host API. They need a representation of what they intended, what a compiler decided on their behalf, what was refused, what was observed after execution and what remains unknown. We present KIR, a typed intermediate representation in which a building is authored as a program held in a versioned repository and lowered to host applications as build targets. KIR is organised around seven ways in which a generator writing into a stateful, partially observable host goes silently wrong, and gives each a representation in data: ambiguous selectors become typed refusals with candidates; defaults keep their provenance; obligations that will not be checked are named before execution; vacuous witnesses are rejected statically; a lost transaction response becomes the state `UNCONFIRMED` with a verify-before-retry rule; the reverse path obeys a census invariant; and decisions are bound by digest to the state they were made against. On a pinned snapshot with 83 operation contracts and Revit as the only backend, offline experiments refuse 29 of 42 stress-test programs with diagnostic codes and no uncaught exception, name 38 of 377 witness obligations as unwitnessable, admit 31 of 100 execution × witness × acceptance combinations under seven stated invariants, and find no vacuous witness in 219 certificate runs; a 60-storey tower is 11,263 characters as KIR against 3,709,235 characters of emitted C#. Native Revit runs are reported from project records and kept separate from reproduced results. A controlled comparison with agents that write host code directly is specified but not yet executed; it is the principal open question.

**a The evaluation building**

— L32

10 m

plan: L32 inside L1 = L60

**60 storeys × 4 m**
**waist 30 % · twist 120°**
generator: 100 lines
3 programs · 6 authored ops
840 expanded ops
780 elements

**b The durable artefact: a typed program**

```
{"op": "stack", "id": "stack2", "levels": 10,
 "h_mm": 4000.0, "base_elev_mm": 120000.0,
 "transform": {"scale_xy_top": [1.057, 1.057],
               "twist_deg_total": 19.07},
 "floor": [
  {"op": "create_floor_by_contour",
   "contour": {"outer": {"shape": "poly",
     "points_mm": [[1381.6, 15337.9], …]}}},
  {"op": "create_column", "id": "column1",
   "xy": [1381.6, 15337.9]}  … 11 more ]}
```

*one of the 6 authored operations (program 2 of 3)*

**expanded for L32 (z = 124 000 mm)**

stack2_L2_floor_by_contour1 —level→ **stack2_L2** create_level
stack2_L2_column1 … column12 —level ×12→

field origins: level, xy MACRO_DERIVED · symbol ENVELOPE_DEFAULT · category REGISTRY_DEFAULT

**c From compiler to host evidence**

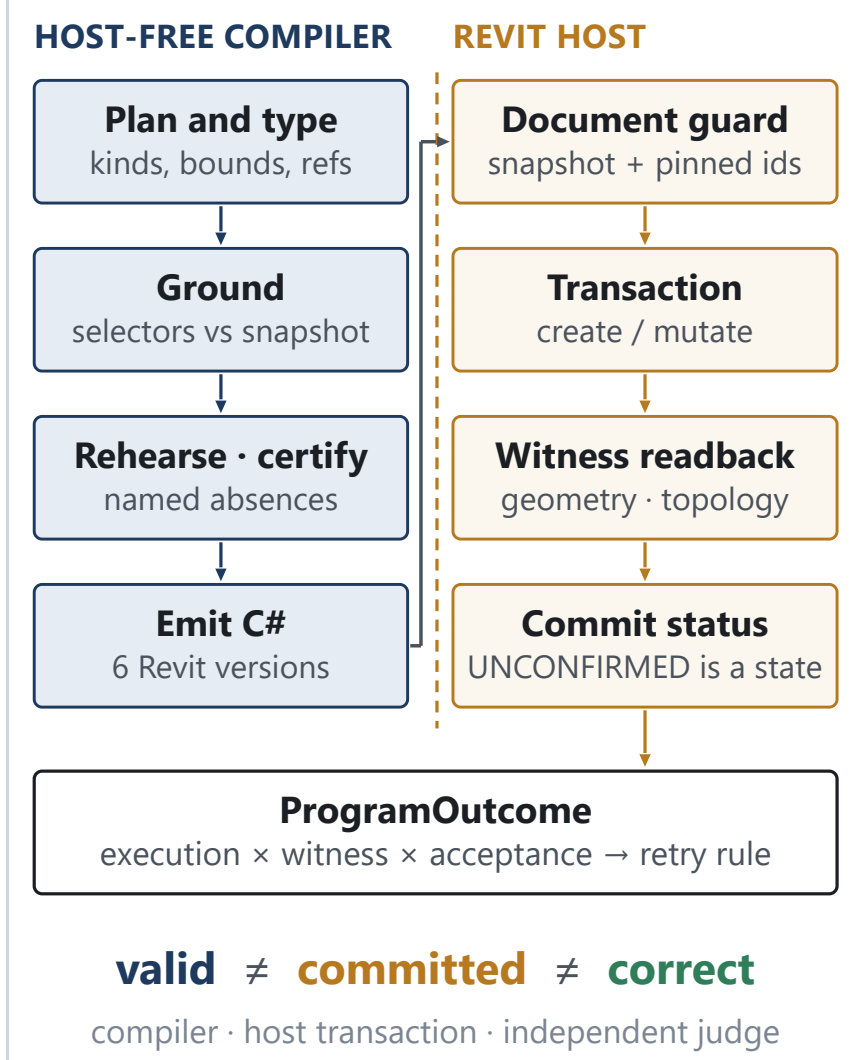

**Figure 1.** The organising idea on the evaluation building. (a) The shipped 60-storey tower, drawn from its own expanded program (60 slab contours, 720 columns; column height drawn as the storey height); in plan, level 32 lies inside the ground and roof contours, which coincide because the 120° twist is four times the 12-gon's 30° symmetry. (b) One of the six authored operations (values rounded) and the typed references its expansion yields, with recorded origins. (c) Compilation, grounding, rehearsal and emission run without the host; creation, readback and commit status are host facts, so "valid", "committed" and "correct" are established by different parties.

## 1 Introduction

Building information modelling (BIM) applications were designed around a human operator. A Revit document is persistent and mutable; its API is in-process and single-threaded; identities and parameter surfaces are host-assigned and release-specific; and the effect of a transaction is only fully known after regeneration and readback [1]. For a person these are properties of the environment. For an autonomous agent they are constraints on reasoning, coordination and recovery: an agent has no use for a viewport, cannot run several API sessions against one document in parallel, and receives from the API a return value or an exception where it needs to know what was decided, what was guessed, what was checked and what was not.

The prevailing practice is to have a language model emit host API code, or call tools that wrap such code. In the product that hosts KIR, 74.0% of 407 tool calls logged between May and June 2026 were free-form generated C#; Autodesk's public Revit MCP server is a shipped instance of the same tool-call pattern [2]. Generated code keeps the agent inside the host's execution model. It does not record why a value was chosen, whether a selector was ambiguous, what was checked after creation, or which facts had become stale before execution. More fundamentally, three claims that are routinely collapsed into one success flag are logically distinct: *the program is valid* (well-typed, references resolved, bounds respected), *the execution committed* (the transaction succeeded and created elements were read back as specified), and *the building satisfies intent* (the model meets the design requirements and applicable rules). KIR keeps these three apart (Figure 1c).

The organising idea is that **a building is a program held in a repository**. Agents author and edit the program; a compiler validates it, resolves its references against a snapshot of the target document, records every choice it makes, rehearses the obligations that will and will not be witnessed, and emits code for a specific host version; execution returns typed receipts and witnesses; authored revisions are persisted with compare-and-swap; and a reverse path reads a document back into the same vocabulary under a census invariant. The host application becomes a build target rather than the agent's workspace. Python remains the language of loops, numerical geometry and generators; KIR is the smaller typed artefact that agents read, diff, merge and compile.

Two concurrent systems share the premise that an agent should produce an inspectable program rather than backend code. CADIR [3] and the CAD-IR component of ArtisanCAD [4] give agents an executable, editable representation over geometric CAD kernels. Their target is construction history and topological editability. KIR asks a narrower systems question for BIM: *what must an agent-facing intermediate representation carry when its target is a persistent, transactional, partially observable document with hosted and levelled elements, host-supplied defaults and outcomes that can be unknown?* Several of KIR's answers have well-known ancestors — plan files and drift detection in infrastructure-as-code, in-doubt transactions, proof-carrying code — and Section 9 draws these connections explicitly. The contribution is to assemble them for BIM, give each a place in one representation, and measure them.

**Contributions.**

- A requirements model for a bureau of agents authoring one building, and a taxonomy of seven silent failure modes that such a representation must make explicit (Section 2, Table 1).
- A typed IR and compiler in which 83 operation contracts declare reference kinds, bounds, field authority, grounding pools, post-conditions and witness tolerances, and whose outputs include machine-readable refusals, per-field provenance and a pre-execution rehearsal (Sections 3–4).
- Execution semantics in which an outcome is a point in a space specified by seven invariants with a derived retry rule, and repository semantics for authored revisions (content addressing, compare-and-swap, typed proposals) plus a reverse path whose losses are counted rather than dropped (Sections 5–6).
- An evaluation on a pinned snapshot that separates reproducible offline measurements from project-reported native evidence, separately re-derives the key counts from the source, and states the unexecuted direct-code baseline as the principal gap (Section 7).

**What this paper does not claim.** It does not show that agents using KIR outperform agents that write C# (that experiment, RQ7, is specified but not run); it does not show backend independence beyond the form of the registry (Revit is the only backend); and it does not claim that a valid, committed and witnessed program describes a good building.

## 2 What a Bureau of Agents Requires

We consider several agents, each responsible for a discipline or a task, authoring one building over an extended period, with a human who reviews and signs the result but does not model. The agents are language-model programs with bounded context, no graphical perception of the model, and the ability to run code. The host is stateful, versioned and only partially observable. Seven requirements follow.

**Compactness:** an agent must be able to re-read the building it is about to change. **Host-free validation:** licensed, serial host round trips are a poor inner loop, so everything decidable from the program and a snapshot must be decided without the host. **Machine-readable refusal:** a refusal must be a record (code, operation, field, expected and observed values, candidates) so that repair is a program rather than a conversation. **Provenance:** every value that reaches the host carries its origin — authored, macro-derived, defaulted or computed. **Concurrent authoring with merge:** revisions, deltas and a three-way merge with typed conflicts. **Coordination queries:** which rooms a wall bounds, what hosts a door, what changed — answered from the representation with an explicit account of what could not be read. **Evidence for the signer:** per element, who decided, on what basis, what was checked and what remains unknown.

Each requirement can be met in form and defeated in substance by one of seven failures that a stateful, partially observable host invites (Table 1). None of the seven is a type error in the usual sense; a program can be well-typed and exhibit all seven. KIR's founding rule, recorded in its specification before the first operation was defined, ranks correctness above diagnosability, diagnosability above coverage, coverage above latency, and latency above convenience, and forbids silent fallback. Refusal is therefore preferred to an apparently successful model whose evidence cannot be reconstructed. The taxonomy was derived from the project's design and defect records rather than from a systematic study of agent failures; whether it captures the failures that matter most in practice is part of what RQ7 would test.

## 3 The Typed Carrier

The vocabulary of Table 1 needs somewhere to live. This section describes the representation that carries it: a small typed program whose operations are contracts, whose references have kinds, whose values carry origins, and whose diagnostics and outcomes are data.

### 3.1 Program envelope and operation registry

A KIR program is a JSON document with an `ir_version`, an optional human-readable `intent`, an `allow_destructive`

**Table 1.** Seven silent failure modes, the representation KIR gives each, and the measurement that exercises it on the snapshot (evidence levels in Section 7).

| Mode | What goes silent | KIR representation | Evidence on the snapshot |
|---|---|---|---|
| **S1** the guess | a selector matches no object or several, and the first is taken | grounding against a stated catalogue; refusal `KIR-G101`–`G104` with the candidate set (§4.2) | 3/3 grounding programs refused in RQ2; refusal record in Figure 2c |
| **S2** the hidden default | an omitted value is filled without record | per-field `FieldOrigin`; a grounding report row for every compiler choice (§4.3) | omitted vs explicit height: identical C#, different plan digests; 5 report rows (Figure 2c) |
| **S3** the unmeasured pass | a check that did not run is counted as passed | tri-state per obligation; rehearsal lists what will not be checked and why (§4.4) | 38 of 377 obligations named unwitnessable; 16 operations witnessed on one axis; 13 of 20 judge rules listed as not evaluated |
| **S4** the vacuous check | a witness exists but cannot fail | witness objects unconstructible without a verdict; static certificate of reachability; axis-honesty test (§4.4) | 0 vacuous results in 219 program-version runs; 511 checks scanned |
| **S5** the collapsed unknown | a lost transaction response is treated as failure or success | `UNCONFIRMED` with retry rule `VERIFY_FIRST`; receipts that cannot claim commit with an unconfirmed error (§5) | 4 of 31 admitted outcomes are verify-first; `StoreCommitUnknown` in the project store |
| **S6** the unaccounted element | a reverse reading drops what it cannot express | typed atoms with reasons; census invariant enforced as a run-stopping error (§6.3) | 92 = 8 + 84 + 0 on a 92-category document; 40 = 16 + 24 at 2, 3 and 4 floors |
| **S7** the drifted premise | a decision correct against a snapshot outlives the state it assumed | plan, contract and grounding digests; document probe and null guards in emitted code; compare-and-swap on the store head (§4.2, §6.1) | one-tolerance change moves the contract digest; stale commit and stale proposal refused (RQ4) |

flag that defaults to false, optional envelope `defaults`, and an ordered list `ops`. Each operation carries two envelope fields besides its parameters: `op`, the discriminator, and `id`, the name by which later operations refer to it. The envelope is closed: an unknown key is refused (`KIR-P003`).

Operations are declared by a frozen registry entry, `OpSpec`, holding the operation family, an ordered tuple of parameter contracts, capability cells, a human-readable post-condition, an effect kind (`READ`, `CREATE`, `MUTATE`, `DELETE`), the kind of reference the operation produces, read/write flags, the parameters grounded against which snapshot pool, per-parameter witness tolerances and a caveat. On the snapshot the registry holds 83 operations: 71 authoring, 6 modify and 6 query; 77 write the model. A content-addressed digest of each contract (`kir-op-contract/2`) lets a compiled plan be bound to the exact contract it was compiled against.

### 3.2 Kinds, units and references

Parameters are typed by a closed vocabulary of 39 kinds — scalar millimetres and degrees, integers, enumerations, points and paths, planes, regions, meshes, surfaces, wall layer stacks, and reference-carrying kinds — enforced when the registry is imported, so that a misspelt kind cannot silently drop a parameter. Length has exactly one public unit, the millimetre; there is no feet kind, and conversion to internal units happens in the emitted code through the host's own unit API. Bounds are declared per parameter, a global coordinate ceiling of 16,000,000 mm applies to all points, and every numeric slot passes a finiteness check, so NaN and infinities are refused rather than coerced.

A reference is an object `{"by": …, "value": …}`: `ref` (a name declared by an earlier operation), `name` and `element_id` (resolved against a snapshot), `default` (only where an operation declares a deterministic rule) and `family_type`. Eight reference kinds (`ELEMENT`, `WALL`, `LEVEL`, `FAMILY_SYMBOL` and four type kinds) are checked between producer and consumer; `ELEMENT` means "any instance", not "anything", a rule narrowed after a live defect in which a move operation accepted a type reference and crashed the host's transform call. Host relationships are thus typed at the registry: a door's `host` slot accepts only `WALL`, so a door hosted by a level is a kind error at planning time, before any snapshot is consulted. Each parameter also carries an *authority* attribute — authored, or derived by the host — and a record of to whom authority passes when the author omits the field; both feed the rehearsal (§4.4).

### 3.3 Provenance and identity

Where an author omits a slot, the value's origin is recorded, not merely filled in. A per-field origin (`FieldOrigin`) distinguishes five cases — explicit, macro-derived, envelope default, registry default and compiler-derived — and a sentinel keeps "omitted" distinct from "written with the same value" all the way to the plan digest. In Figure 1b, each expanded column records that its `level` and `xy` were supplied by the `stack` macro (`MACRO_DERIVED`), its family symbol by the program's envelope defaults (`ENVELOPE_DEFAULT`) and its category by the registry (`REGISTRY_DEFAULT`). Operation identity is a string of one to sixty-four characters, unique within the program (`KIR-P006` on duplicates); the DSL assigns per-kind counters when the author writes none. Host element identity is a separate concept that the paper does not rely on.

### 3.4 Generated front ends and a worked example

Neither Python front end contains a hand-written builder. The SDK and the DSL construct one function per registry entry at import time, so their operation-name sets equal the registry's (83 = 83 = 83); the JSON Schema is generated from the same registry (86 entries: the 83 operations and the macros `stack`, `grid_array` and `series`). Forty-four non-test modules import the registry. The emitters are the measured exception (§7.1).

Figure 2 follows the README room through the toolchain. The loop and the handles are Python; the five operations, the level selector and the typed host reference are KIR. Compiled against a snapshot whose levels are named “Ground Floor” and “First Floor”, the program is refused with a record naming the operation, the field, the value that did not resolve and the candidates that were searched; after the selector is corrected, the compiler emits 37,160 characters of C# for Revit 2026 and returns, as data, the five choices it made on the author's behalf. At building scale the same pattern holds: the evaluation tower of Figure 1 is generated by 100 lines of Python into three programs of two authored `stack` operations each, which expand to 840 operations and 780 elements. The analogy with a compiler IR is deliberate and bounded: a machine target's semantics are known to the compiler, whereas a BIM host can be observed only in part, so the IR must carry evidence states and not only types.

## 4 Compiler: Refuse, Record, Rehearse

The single entry point, `compile_program`, performs a deterministic host-free pass before emission and never raises: an unexpected internal error becomes `KIR-P000` with an incident id. Figure 3 shows its stages, the diagnostic families each can raise, and the boundary between what is established statically and what only the host can establish.

### 4.1 Plan and reference discipline

Planning parses the envelope, expands macros, applies envelope defaults, validates each operation against its contract — kinds, bounds, closed enumerations, finiteness, unknown fields refused — rejects duplicate ids and mixed query/write families, and checks references: every `ref` and every relational address must name an operation strictly earlier in program order (`KIR-L003`) whose result kind the consuming slot accepts (`KIR-L004`). The output is an immutable, canonically hashed `PlannedProgram` with per-field provenance and a whole-plan digest.

Two stages a reader might expect are absent by design. There is no dependency graph as a data structure: the author's order is the schedule and only backward references are legal, so cycles are inexpressible and no cycle diagnostic exists. There is no transaction-planning stage: emission uses one transaction per program or one subtransaction per operation, and a byte-budget chunker that splits large programs into contiguous slices records in its plan that atomicity is thereby weakened to per-chunk.

### 4.2 Grounding: refuse the guess, bind the premise

For writing programs a `GroundingContext` binds a content digest of the supplied snapshot to a document, profile and revision. Name, default and family/type selectors are then resolved against the snapshot's pools. A required pool with no snapshot is `KIR-G103`; a name with no match is `KIR-G101` with the candidates searched; several matches is `KIR-G102`; an empty pool is `KIR-G104`. This is the mechanism for S1, and its refusal is a record rather than a message (Figure 2c). An `element_id` selector is the deliberate exception: it is passed to a null guard in the emitted code rather than checked against the offline snapshot, and RQ2 records it as the one reference-error class that is not detected offline. Relational addresses — a point at the intersection of two grids — are resolved to literal millimetre points here, and geometry checks that need grounded numbers, such as a zero-length wall or a hosted opening beyond its wall, run after grounding.

The binding against S7 has four parts. The plan digest changes when any authored operation changes; the operation-contract digest changes when any declared field of a contract changes (changing one wall tolerance from 5.0 to 5.5 mm moves it from `a13f5495…` to `451eb4fa…`, and restoring the value restores it); the grounding-context digest changes with the snapshot or the document it describes; and the emitted code re-checks the premise inside the transaction, immediately before the effect, through a document probe and a null guard per grounded id (§5.2).

### 4.3 Record: the receipt of named defaults

Provenance is a hashed fact about how a value was decided, not about what it is. On one wall, omitting `height_mm` gives the origin `REGISTRY_DEFAULT` and writing `height_mm: 3000` gives `EXPLICIT`; the resolved value is the same 3000.0, the emitted C# is byte-identical and the contract digest is unchanged, yet the plan digests differ (`f1f8fba9…` against `6f78ebf8…`). Choices the compiler makes during grounding — which level, which type, which symbol, by which rule — are returned as rows of a grounding report. The design intent is recorded in the source: a choice that cannot be shown to the author is a first-match lookup under another name. Where the host rather than the compiler will decide, the row records an address from which the host's choice will be read back (Figure 2c) instead of a value the compiler invented.

### 4.4 Rehearse and certify: name the unmeasured pass and the vacuous check

Two static modules act before any host is involved. The *rehearsal* takes a program and reports, without building anything, the obligations that will be checked per operation and axis (geometry, topology, semantic, parameter, identity), the values the host will overwrite regardless of what the author wrote, the fields whose omission transfers authority, the operations witnessed on one axis only, and the *named absences* — obligations the project has decided not to witness, each with its reason. On the snapshot the obligation table declares 377 clauses over the 77 writing operations (from 2 to 13 per operation), and 38 clauses over 26 operations are named absences: 27 exemptions (pre-flight refusals, host-derived values, emission-order rules), 8 absences proper — for example, rooms bounded by a moved wall are not re-checked because no reverse index exists — and 3 caveats. The rehearsal is the mechanism for S3: a program's owner learns, before spending a host round trip, which parts of “the build succeeded” will be a measurement and which an assumption.

The *certificate* is a static analysis of the emitter's own C#. For each operation and version it checks that the creation call is guarded by a typed refusal and that every post-condition in the operation's obligation table has a reachable marker; a certificate whose verdict statement is dead code is vacuous and fails. Two further guards act at construction and test time: a witness object cannot be constructed without a verdict statement, and a static test parses the C# of 511 witness checks across 68 operations and fails any geometry-tagged check that reads back only a parameter the emitter itself wrote, unless it is on an eight-entry allow-list with a stated reason — a rule adopted after exactly that defect in the wall emitter. These are the mechanisms for S4. The certificate is disabled by default in the **a Python authoring (kir.dsl)**

```python
from kir import dsl
dsl.reset(intent="4 walls + 1 door")
level = dsl.by_name("Level 1")
corners = [(0, 0), (6000, 0),
           (6000, 6000), (0, 6000)]
walls = [dsl.create_wall(p0_mm=a, p1_mm=b,
           level=level, height_mm=3000.0)
         for a, b in zip(corners,
             corners[1:] + corners[:1])]
door = dsl.create_door(host=walls[0],
                       offset_mm=2500.0)
program = dsl.build()
```

Python keeps the loop and the handles;
the program it builds is five typed operations.

**b The KIR program (5 operations)**

```
{"ir_version": "1.0",
 "intent": "4 walls + 1 door",
 "ops": [
  {"op": "create_wall", "id": "wall1",          ❶
   "p0_mm": [0, 0], "p1_mm": [6000, 0],
   "level": {"by": "name",                       ❷
             "value": "Level 1"},
   "height_mm": 3000.0},                         ❸
  … wall2, wall3, wall4 …
  {"op": "create_door", "id": "door1",
   "host": {"by": "ref", "value": "wall1"},      ❹
   "offset_mm": 2500.0}]}
```

❶ id the author did not write (DSL counter)
❷ grounded selector: resolved against a snapshot
❸ one public length unit: millimetres
❹ typed reference: kind WALL, the only kind accepted

**c What the compiler returns**

**refused · KIR-G101**
op `wall1` field `level` got `"Level 1"`
candidates 43 `"First Floor"`, 42 `"Ground Floor"`
blame: author · no C# is emitted
*(snapshot with two levels, no host)*

↓ *selector changed to "Ground Floor"*

**accepted · 37 160 chars of C#**
receipt of named defaults:
`wall1–4 type doc_default`
chosen by Revit, read back from
`result.wallN.type_name`
`door1 symbol sole_entry` → id 700

**rehearsal · 5 ops · 0 host trips**
18 obligations checked; 8 not:
top_level omitted → the wall top is
a number, no longer the level

**Figure 2.** A worked example, reproduced with KIR 0.8.1. (a) The README room written with the DSL. (b) The program it lowers to. (c) The compiler's replies without a host, against a test snapshot with two levels and one door symbol: a refusal record for the unresolved selector (S1); after the fix, C# for Revit 2026 and the receipt of named defaults — four wall types left to the host and read back by address, one door symbol chosen as the sole pool entry (S2); and the rehearsal, which reports before any host call that omitting `top_level` transfers authority and removes eight obligations (S3).

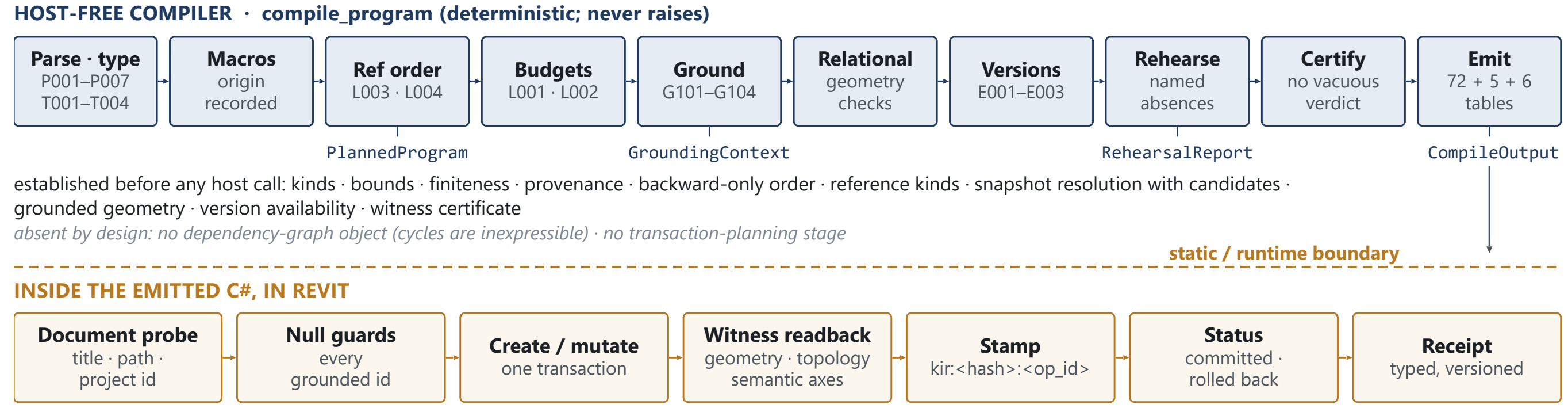

**Figure 3.** The compiler as implemented, the diagnostics each stage can raise, and the static/runtime boundary. Everything above the dashed line is decided from the program and a snapshot; everything below runs inside the emitted C# in the host.

serving path, so on the snapshot it is a test-time guard, not a runtime one.

### 4.5 Emission

Emission dispatches each operation through three hand-maintained tables (72 + 5 + 6 = 83) and concatenates declaration, creation, post-condition and readback fragments into one transaction body. Version branching is inline per emitter (ceilings refused before Revit 2022; the floor-creation overload changes in 2022; 64-bit element ids from 2024), and a length is wrapped in a conversion to internal units that delegates to the host's unit API for the exact version in use — so a unit defect in an emitter is a runtime defect, discoverable only by execution. Compilation of the emitted text by Roslyn against the real Revit API assemblies is performed by an external service; the repository's offline gate is offline with respect to Revit, not to Roslyn.

## 5 Execution: Commit Is Not Correctness

Everything up to emission is deterministic Python: the same program and snapshot yield the same digest and the same text. Everything from dispatch onward happens in a host the compiler does not control. KIR's contribution on that side is a vocabulary in which the host's facts, including the fact that a fact is unknown, are reported.

### 5.1 The outcome space

An outcome is a triple of execution state (`NOT_STARTED`, `READ_COMPLETED`, `COMMITTED`, `ROLLED_BACK`, `UNCONFIRMED`), witness state (`NOT_RUN`, `SATISFIED`, `VIOLATED`, `INCOMPLETE`) and acceptance state (`NOT_APPLICABLE`, `NOT_RUN`, `ACCEPTED`, `REJECTED`, `INCONCLUSIVE`), where acceptance is the verdict of an independent post-execution pass. The implementation calls this an outcome algebra; it has no algebraic operations, and we describe it as what it is — a product type whose constructor enforces seven invariants (Figure 4):

**I1** a terminal witness (`SATISFIED`, `VIOLATED`) requires a completed execution (`READ_COMPLETED`, `COMMITTED` or `ROLLED_BACK`);

**I2** `NOT_STARTED` ⇒ witness `NOT_RUN`; **I3** `UNCONFIRMED` ⇒ witness ∈ {`NOT_RUN`, `INCOMPLETE`};

**I4** `READ_COMPLETED` ⇒ acceptance `NOT_APPLICABLE`; **I5** acceptance `NOT_APPLICABLE` ⇒ `READ_COMPLETED`;

**I6** acceptance ∈ {`ACCEPTED`, `REJECTED`} ⇒ `COMMITTED`; **I7** `ACCEPTED` ⇒ witness `SATISFIED`.

Of the 100 syntactically possible triples, 31 are admitted. Attributing each of the 69 refusals to the first invariant it violates, in constructor order, gives 20, 5, 0, 16, 11, 14 and 3: I3 is never the first to fire because I1 already excludes a terminal witness on an unconfirmed execution. The retry rule is derived, not stored: a committed outcome forbids retry, an unconfirmed one requires verification first, and every other admitted outcome is safe to retry (14, 13 and 4 outcomes respectively). One consequence is visible at a glance: `ACCEPTED` is reachable from exactly one cell, a confirmed commit with a satisfied witness. The factory for a lost response yields `UNCONFIRMED` with witness `INCOMPLETE` and acceptance `NOT_RUN` — the representation of S5. Figure 4 is generated by enumerating the constructor itself, not drawn by hand. The invariants constrain the representation; they are not a proof about the host, whose behaviour the paper measures only through project records.

### 5.2 Transactions, document binding and stamps

Emitted code runs one transaction per program, or one sub-transaction per operation, which forces post-conditions into report mode because whole-program rollback is no longer possible. A failures preprocessor suppresses the host's warning dialogs so that a warning cannot silently cancel a run. Before any write, a document probe compares the live document's title, path and project identifier with the values the program was grounded against, and returns a document-mismatch refusal on divergence; every grounded id is re-fetched and null-checked, and a null rolls the transaction back as “model changed after grounding”. Each created element is stamped in the same transaction with `kir:<program-hash>:<op_id>`. Resumption after a failure is at chunk granularity, keyed by a content hash of the chunk, and a journaled map from chunk to created element ids is checked against the plan before anything is reused. One bound is recorded as open: the stamp is written after creation, so it cannot by itself exclude a crash between creation and stamping.

### 5.3 Receipts and witnesses

A commit receipt is a typed, versioned record of run id, operation, element ids, bridge error, commit confirmation, status, program id and document revision. Its constructor refuses a receipt that claims a committed status while carrying a bridge error without confirmation, on the ground that such an outcome is unknown rather than negative. The raw facts originate in a C# connector that keeps an fsynced operation journal keyed by a client-supplied operation identity and, on a repeated request, returns the cached receipt or a typed “running, unknown” refusal. A witness is an obligation read back from the created element inside the emitted code; a check signs the geometry, topology or semantic axis only if it read that axis, and a separate tri-state records whether an axis was declared at all, so that an unchecked axis is distinguished from a satisfied one. A violated post-condition before commit rolls back and is reported as `KIR-X004`; one observed after commit is `KIR-W004`, because reusing the former would wrongly imply that a retry is safe.

### 5.4 Intent verification

Readback exists at three distances from the operation: the in-transaction witness; a post-write census probe compared with a program-derived expectation; and an independent re-extraction of only the receipted elements, from which a spatial model is built and judged by the same habitability rules as the program's own claims, with the two sources typed apart (`PROGRAM` versus `PARSE`). This is how completion of execution and satisfaction of intent become two measurements. The judge carries twenty habitability rules and reports, beside its findings, the rules it could not evaluate and why; on the demonstration house and on the README example, 7 of 20 rules are evaluated and 13 are listed as not evaluated with a reason each. Coverage the judge did not have is not reported as a pass.

## 6 Persistence and the Reverse Path

A building held as a repository needs two histories: what the agents authored and what the host holds. KIR keeps both, in two owners that reference each other by address and are deliberately not merged.

### 6.1 Authored revisions and compare-and-swap

The unit of authorship is a module instance whose named outputs have an identity independent of position and content, so an output keeps its id when unrelated outputs are inserted or reordered — the property an agent's second edit needs. A revision is immutable and content-addressed: its id is the SHA-256 of the canonical body. Persistence is one SQLite file per project with an additive schema ladder: a store from an earlier release opens under a later one and refuses features that need an upgrade instead of silently degrading. Commit is compare-and-swap: a single conditional update that moves the head only if it still equals the expected revision, and refuses with `StoreConflict` when it does not. A commit whose acknowledgement itself fails is `StoreCommitUnknown` — S5 at the level of the store — and is recovered by retrying the identical revision, which the store recognises by id and byte-compares without moving the head. The store's own documentation records that its durability tests simulate process crashes, not power loss.

### 6.2 Proposals, merge and diff

An agent need not commit directly; it proposes. A change proposal carries its base revision, a candidate child, a declared scope of instances and fields it may touch, an author and a reason, and its constructor checks that its own diff stays inside the declared scope, so a proposal cannot grant itself authority by widening its serialised scope. Acceptance refuses unless the store's head is the expected one and the proposal's base is a stored ancestor, then performs a three-way merge and commits under the store's own compare-and-swap, so acceptance is guarded twice. Divergent edits that do not overlap, computed through the diff's dependency closures, are rebased; overlapping edits or incomplete dependency analysis return typed conflicts and commit nothing. The merge runs no compiler, geometry or native code and says so in its report. The diff reports by address — added, removed, changed and affected outputs, each with a reason — and reports a moved output as re-keyed only when the pairing is unambiguous. Figure 5a shows the RQ4 exercise.

### 6.3 The reverse path and the census invariant

Reverse reading passes through extraction, lifting and folding. Each extracted element must lift to a typed operation or to an *atom*: a node that says “this element exists, and here is the typed reason it could not be expressed” — a registry gap, missing geometry, metadata, parameter or reference, an unsupported curve kind, an unresolved dependency, a generator child, or a placement state that could not be read. The invariant binding these stages is enforced as a run-stopping error: at extraction, the count the host reports must equal extracted plus explicitly unscanned; at verification, lifted leaves must equal operations plus atoms. On the synthetic 92-category document, lifting yields 8 operations and 84 atoms with eight distinct reasons and nothing unread (Figure 5b); a chain test asserts 40 = 16 + 24 and re-asserts it at three and four floors. Four sibling laws have code and passing tests: every side-stage extraction returns rows plus failures; a witness may sign only the axis it read; a partial read of a work-shared document marks everything derived from it — adopted after a real document in which seventeen closed worksets silently yielded eleven elements instead of two thousand and sixteen; and the shipped compiler carries no device identifier or foreign path. The central claim is not that reverse coverage is complete; it is that missing coverage cannot silently disappear.

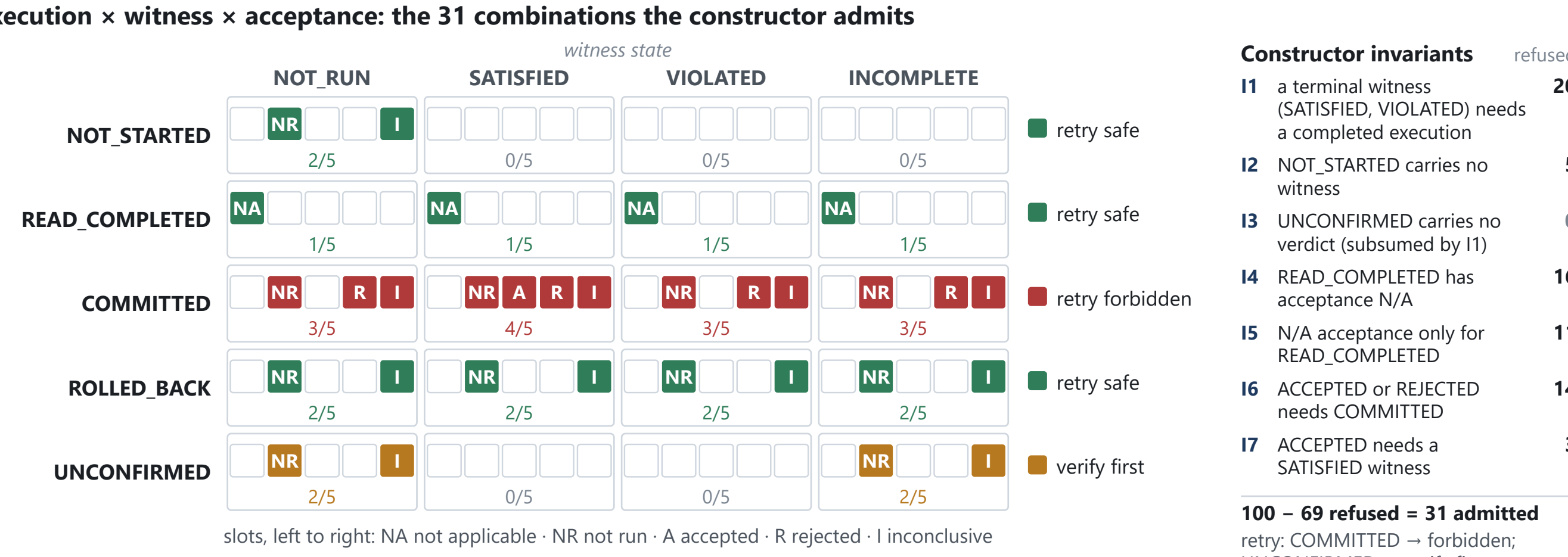

**Figure 4.** The outcome space, computed by enumerating the `ProgramOutcome` constructor of KIR 0.8.1 over all 100 triples. A filled slot is an admitted acceptance state; colour is the derived retry rule. Right: the seven constructor invariants and the number of refused triples first rejected by each.

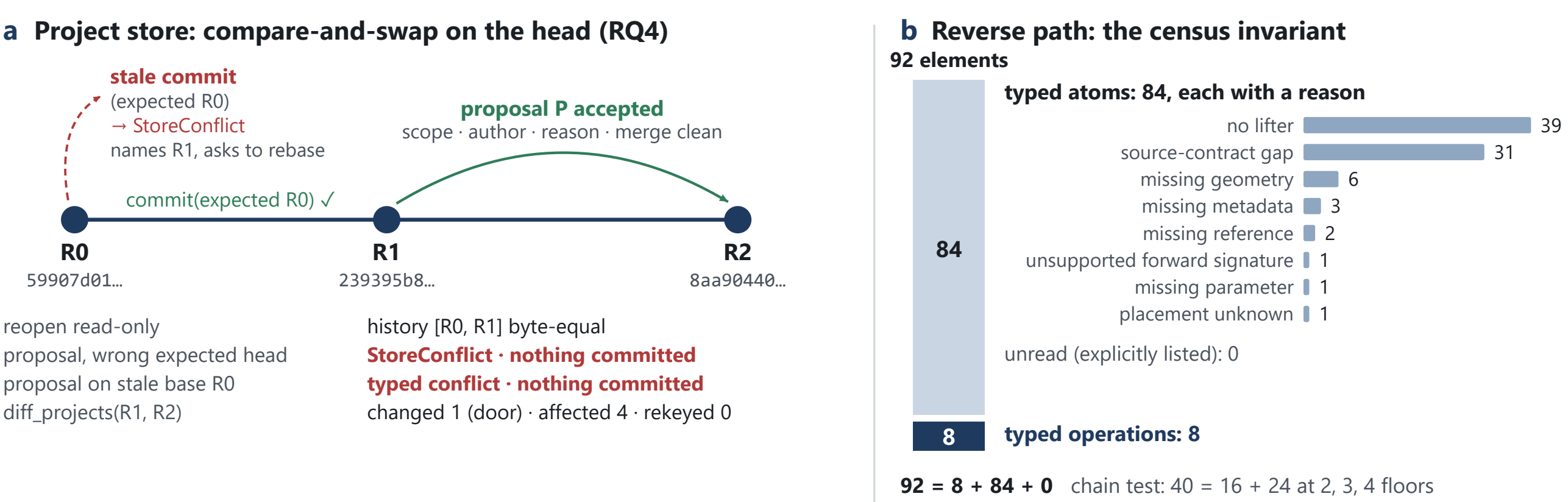

**Figure 5.** The two persistence invariants, as exercised offline on the snapshot. (a) Authored revisions commit by compare-and-swap: a stale commit, a proposal with the wrong expected head and a proposal on a stale base are refused or returned as typed conflicts with nothing committed. (b) The reverse path on a synthetic document with one element in each of 92 extractable categories: every element becomes a typed operation, a typed atom with one of eight reasons, or an explicitly listed unread item.

The observed building is versioned separately, by an append-only, hash-chained journal of canonical operations from which identity-bearing fields are removed, so two revisions compare as multisets and merge three-way; a delta materialises only retirements and emissions and reuses untouched elements. Keeping authored identity in the store and observed state in the journal is a design decision: the journal's identity-free form would make two authors' identical new elements indistinguishable.

## 7 Evaluation

All offline measurements use the source tree of KIR 0.8.1 with Python 3.12.13 on Linux 5.15, 8 logical CPUs and 15 GiB of memory, with no Revit, bridge or network. The optional OCCT geometry kernel was absent, so kernel-gated tests skip by name rather than pass. We label evidence by level: IMPL (source-level implementation), TEST-OFF (offline tests or fixtures), EMIT (present in emitted code), ROSLYN (compiled against Revit API assemblies; project-reported), LIVE (native Revit execution; project-reported) and FUTURE. Table 2 summarises the research questions. For this revision the key counts and the RQ3 emissions were also re-derived from a second source archive of the same release in another environment — a cross-environment check, not a third-party replication; the structural counts agree exactly (Supplement Table S9).

### 7.1 RQ1 — Representation consistency

Registry, SDK builders and DSL table have equal operation-name sets (83 each); no operation shows a missing, extra or renamed parameter in its builder signature, schema entry or documentation. This holds largely by construction, because each surface iterates the registry, and we report it as a property of the architecture rather than as an empirical finding. Emission does not hold by construction: the 77 writing emitters contain 297 string-literal slot accesses and none through the registry, so a slot renamed in the registry is caught by tests and golden files, not by the type of the code. Compiling one minimal program per operation for each of six Revit versions makes 498 operation-version attempts (2.38 s): 482 emit and 16 fail closed at explicit version gates across seven operations, each as a typed refusal (`KIR-E003`); 76 operations emit for all six versions, 3 for five, 3 for three, 1 for two.

### 7.2 RQ2 — Rejection of invalid programs

The corpus consists of a valid baseline — a level, four walls, a door and a window hosted on two of them, and a room — against a snapshot with one door symbol, one window symbol and one level, plus 41 variants in eight categories, each written after reading the enforcement code, with its expected behaviour recorded before the run. All 42 programs returned without an uncaught exception (Figure 6a). Twenty-nine were refused with a KIR code; 27 of these name both the operation and the field, and the other two are envelope-level and name the field only. Thirteen were accepted: ten as intended, and three silent acceptances of a real defect, which are the principal finding. A family symbol selected by `element_id` and absent from the snapshot compiles to code whose null guard refuses at execution — a documented design decision, so this reference-error class is not detected offline. A wall height and a level elevation of 3.0, almost certainly metres, are legal three-millimetre values, because the unit discipline is a type discipline without a plausibility check. One case classified in advance as legal was refused (an extra envelope key), and one geometric case was caught by a different law than intended: a symmetric bow-tie floor has near-zero area and was refused as degenerate before the self-intersection detector ran. Reference cycles cannot be formed at all.

### 7.3 RQ3 — Version separation and size

Five programs — the README's four-wall room, the same room with a door, and the tower's three programs — emitted for six Revit versions give 30 emissions, all without diagnostics; three emissions of the room for Revit 2026 are byte-identical. Version-specific text appears exactly where the emitters carry version knowledge: the room's texts for 2021–2023 are identical and differ from 2024 onward by the migration from `ElementId.IntegerValue` to the 64-bit `ElementId.Value`; the tower's 2021 text additionally differs in the floor-creation overload.

For Revit 2023 the room is 612 characters of compact KIR against 34,420 of C#, and the tower 11,263 against 3,709,235, a ratio of 329 (Figure 6b). This ratio is easy to over-read, and we state what it measures. It is the expansion of lowering — per-element guards, transaction scaffolding, identity capture and witness code — and therefore the cost an agent would pay if its working copy of the building were the emitted code. It is *not* a comparison with the C# an agent would write by hand: a hand-written generator for this tower would be a loop of roughly the size of the 100-line Python script that produces the three programs. The argument for KIR is rather that it is the smallest artefact we know of that carries operation identity, reference kinds, per-field origin and contract digests, and can be compiled, diffed and merged without re-running the generator.

### 7.4 RQ4 — Persistent evolution

Two exercises use public APIs on temporary stores (Figure 5a). In the project store, R0 is created, R1 is committed against the correct head, a stale commit is refused with a conflict that names the current head and asks for a rebase, the store is reopened read-only with both revisions byte-equal, a proposal is accepted as R2, a proposal with the wrong expected head is refused, and a proposal on a stale base is returned as a conflict with nothing committed; the diff of R1 and R2 names one changed output (the door) and four affected ones. In the journal of observed state, three revisions of a small building reuse four of five and four of six leaves, serialise to 2,919 bytes and reload with the chain verified; a three-way merge reports one modify/modify conflict on a shared wall and none on an independent wall, where the clean merge equals the direct application of both edits. The corresponding test files pass (233 tests with 31 kernel-gated skips, and 186 tests).

### 7.5 RQ5 — The evidence mechanisms, measured

*Provenance and binding.* Omitted versus explicit defaults give the same value, byte-identical C#, the same contract digest and different plan digests; a one-tolerance mutation changes the contract digest and its restoration restores it; the demonstration house's grounding report has five rows. *Named absences.* The obligation table declares 377 clauses; 38 are named absences over 26 operations. Over one minimal program per operation, the rehearsal finds 3 operations with no witnessed axis (queries), 16 with one, 31 with two, 27 with three and 3 with four; three query operations appear in no corpus. *Certificates.* Over the 73-program golden corpus and three versions, none of the 219 program-version runs is vacuous (per-version counts in Supplement Table S4). *Outcome space.* Exhaustive enumeration admits 31 of 100 triples (§5.1). *Census.* Eleven test files pass (130 tests, two environment-gated skips), and each of the five conservation laws has enforcing code and a passing test. *Judge and instruments.* The judge evaluates 7 of 20 rules on the demonstration house and lists 13 as not evaluated with reasons; of thirteen shipped instruments run on a bare installation, seven return a number, three a typed refusal naming the missing environment and three print usage — none returns an unmeasured zero.

### 7.6 RQ6 — Native execution (project-reported)

The records below are the project's operational runs, held as artefacts outside the repository and cited with date, host version and denominator; we did not re-run them (Figure 7). On 2026-07-27, against Revit 2023 and a production model, 20 writing operations were executed through the bridge: 18 built with their witnesses satisfied, one was broken and later repaired, and one had no sub-strate in that model; with eight operations proven live earlier, 26 of 28 named operations had built live by that date. On the same date the project's compile service compiled 207 programs against the real Revit API assemblies for all six versions — 1,242 checks with no final failure, after an interim run exposed 16 failures traced to a version-tagging defect. On 2026-09-03, against Revit 2026, 21 operations were executed with per-operation witnesses: 17 built, two had nothing to build on because their type pools were absent from the document, one failed a precondition (a room separator needs a floor-plan view) and one returned a bridge error without text. A first attempt reported 20 operations as broken because of name collisions with earlier fixtures, a defect of the instrument that the project corrected — S3 at the level of the measuring instrument. The reverse path's coverage of 92.83% over a 90,758-element model is an offline re-lift over a live capture. The only agent-in-the-loop evidence is three bureau runs (§8) with two external language-model responders, reported in project notes as passing 9 of 9 checks; these show that the exchange works, not that agents succeed more often. Numbers that appear in the project's README without an artefact are not used.

**Table 2.** Research questions and primary results on KIR 0.8.1. RQ1–RQ5 are reproducible without Revit; RQ6 is project-reported; RQ7 is specified but not executed. Full denominators are in the supplement.

| Question | Experiment | Result | Evidence |
|---|---|---|---|
| RQ1 consistency | registry vs generated surfaces; one program per operation × 6 versions | 0 parameter mismatches over 83 operations; 498 attempts: 482 emitted, 16 typed version refusals; 297 literal slot accesses in emitters | IMPL, TEST-OFF |
| RQ2 rejection | 41 stress variants + a valid baseline | 29/42 refused with a code, 27 naming operation and field; 0 uncaught exceptions; 3 silent defects accepted | TEST-OFF |
| RQ3 versions, size | 5 programs × 6 versions; repeated emission; sizes | 30/30 emissions; byte-identical repeats; tower 11,263 vs 3,709,235 characters (329×) | IMPL |
| RQ4 persistence | R0→R1→R2, stale commits, reopen, two merges | stale head refused; history reopens byte-equal; one conflicting and one clean merge; untouched leaves reused | TEST-OFF |
| RQ5 evidence | obligations, certificate, outcome space, census, judge | 38/377 named absences; 0/219 vacuous; 31/100 admitted; 92 = 8 + 84 + 0; 13/20 rules listed as not evaluated | IMPL, TEST-OFF |
| RQ6 native host | project operational records | Revit 2023: 26/28 named operations built by 2026-07-27; Revit 2026: 17/21 on 2026-09-03, four exceptions classified; 1,242/1,242 compilations | LIVE, ROSLYN |
| RQ7 baseline | same model and budget: direct C# vs `kir.dsl` | protocol specified; not executed | FUTURE |

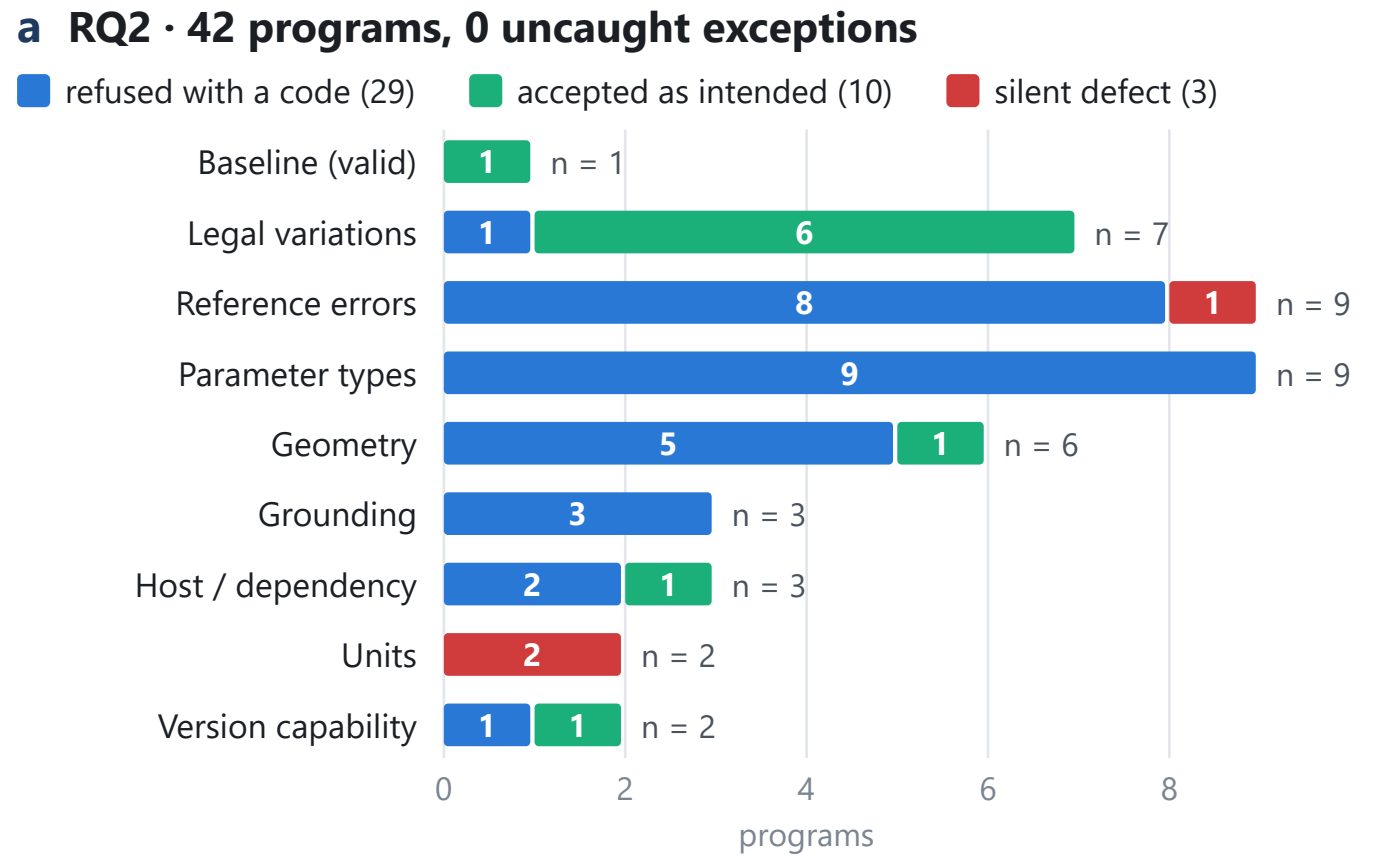

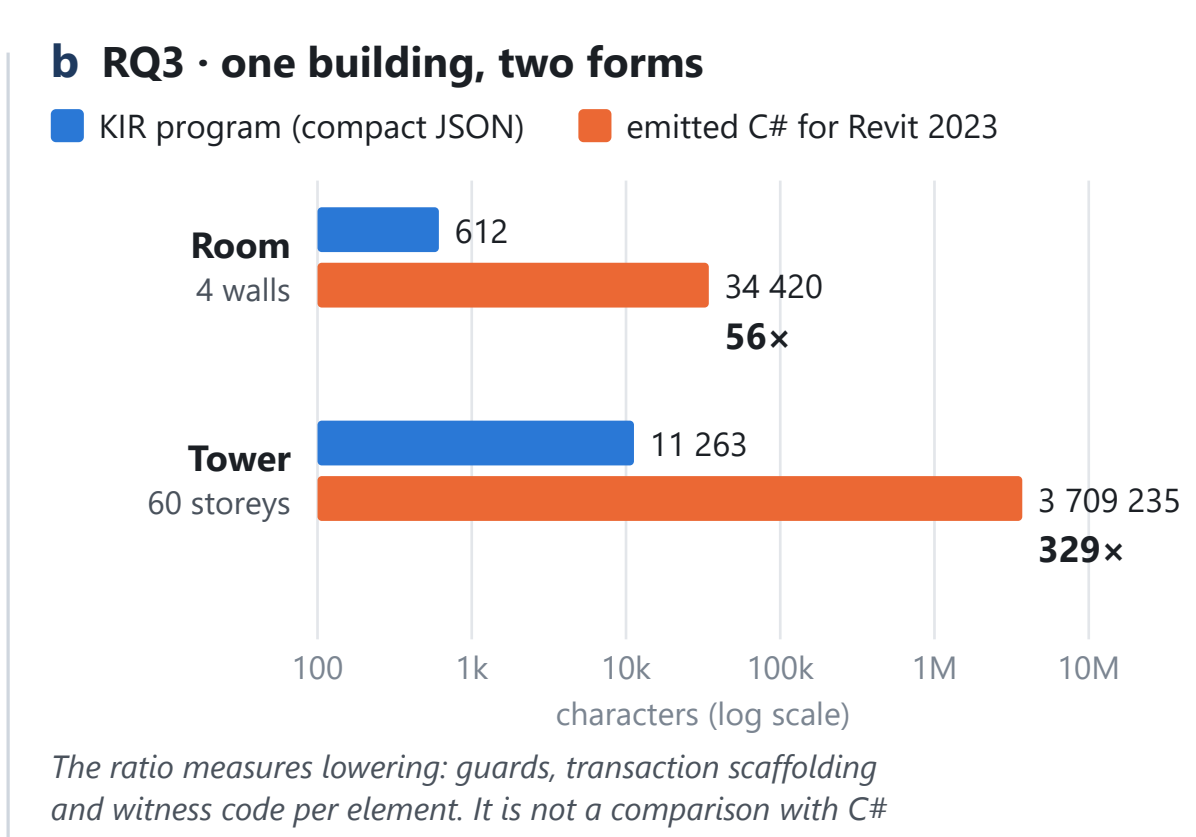

**Figure 6.** Two primary measurements. (a) RQ2: outcome of each of the 42 corpus programs by category; the three silent defects are an `element_id` selector deliberately deferred to a runtime guard and two metre-intended lengths accepted as millimetres. (b) RQ3: the same building as a compact KIR program and as emitted C# for Revit 2023, on a log scale. The table view of (a) is Supplement Table S6.

### 7.7 RQ7 — The missing baseline

The principal empirical gap is exposed rather than hidden. RQ7 compares the same language model under the same prompt budget in two arms — writing Revit API C# directly, and writing Python against `kir.dsl` — on ten tasks of increasing statefulness, from a single room to an edit of a hosted door after its wall moved and a conflicting concurrent edit. Scoring predicates are fixed before the run: Roslyn compilation, native commit, post-condition readback against stated intent, preservation of untouched identities on edit tasks, repair rounds, the ability to show effects before execution, and document state after a forced unknown outcome with the bridge killed mid-transaction; a KIR compile-time refusal counts as a failure unless repaired within the same budget. At least two Revit versions are required. The full protocol is in Supplement Section S5. Until it is run, KIR's argument against direct code is mechanistic rather than a controlled head-to-head result.

### 7.8 Threats to validity

*Construct.* Refusal counts toward rejection only where refusal was the recorded expectation; compactness measures lowering, not an alternative authoring practice. *Internal.* The RQ2 corpus and most tests were written by the project after reading the enforcement code, which risks fitting **a Evidence ladder on the snapshot**

|  |  |
|---|---|
| **Intent verified** PROGRAM vs PARSE verdict | comparator offline |
| **Observed** witness · census · re-extraction | 92.83 % re-lift of 90 758 |
| **Committed** transaction status · receipt | 26/28 (2023) · 17/21 (2026) |
| **Dispatched** connector · operation journal | contract + fixtures |
| **Compiled** Roslyn vs Revit API assemblies | 1 242 / 1 242 |
| **Emitted** C# text · golden corpus | reproduced offline |

reproduced here · project-reported (artefact) · contract + fixture tests

**b Native Revit runs (project-reported)**

**Revit 2023 · 2026-07-27 · production model** 26 / 28
failure: create_dimension (later repaired)
no substrate: create_beam

**Revit 2026 · 2026-09-03 · document "1"** 17 / 21
no type pool: ceilings, railings · precondition: room
separator needs a plan view · failure: opening by host face

**Compile service · 2026-07-27**
207 programs × 6 versions = 1 242 checks, 0 final failures
(16 interim failures: a version-tagging defect, fixed)

built + witness · proven earlier · nothing to build on · precondition · failure tests to the implementation; expectations were recorded before each run, and the three silent acceptances were reported, not repaired. *External.* One host (Revit), one project's traffic, and synthetic documents for the census. *Independence.* The project's internal audits were performed by language-model agents under the project owner; no external replication or user study exists.

**Figure 7.** Evidence is a ladder, not a Boolean. (a) Each rung, the artefact that represents it, and the level reached on the snapshot: reproduced in this paper, project-reported with an artefact, or present as a contract with fixture-driven tests. (b) The project's native runs by date and host version, with every exception classified.

## 8 Coordination Built on the Representation

Once the building is a versioned typed program, coordination can be expressed as repository operations over inspectable state rather than as a transcript of messages. The repository builds several such mechanisms on the foundation of Sections 3–6; they are implemented and tested on the snapshot (Figure 8; Supplement Section S4). A *building graph* (Figure 8a) lifts an extracted document into typed nodes and eleven kinds of relation — hosted in, on level, bounds room, joined to and others — each with a modality (proven, possible, refuted, unresolved); an edge whose host is absent is kept as unresolved instead of being dropped, and relations whose sources were not supplied are listed as not measured. *Clash detection* (Figure 8b) builds conservative hulls under a census and, when the OCCT kernel is present, an exact narrow phase; one of its refusals was added by measurement after the kernel's Boolean intersection was found to return zero volume while reporting success for twisted lofts of nine degrees or more — exactly where the project's demonstration towers sit. *Refinement* records lineage from conceptual parts to detailed elements with residue and deviation as addressed numbers; captured buildings can be edited offline with every field marked represented, approximate, source data or unknown. A *habitability judge* carries thirty thresholds, eighteen of them jurisdictional: the same 2,400 mm room is flagged under a 2,500 mm floor and passes under a 2,286 mm floor.

The *bureau* runtime coordinates several workers over one project store: a mission is a plan of turns with backward-only dependencies; two workers claim steps under the store's compare-and-swap, so the loser takes the next free step; a conflict returns as a replan request naming the revisions and affected outputs, and a second conflict — or three turns without progress — stops the team by name (85 tests; Supplement Figure S1). The drawing-recognition pipeline for which KIR was conceived becomes one authoring agent among others, subject to the same refusal, provenance, merge and witness machinery.

## 9 Related Work

**Direct host automation.** The Revit API [1], RevitPythonShell [5], pyRevit [6] and Dynamo [7] are imperative interfaces to one host at one version: identity is the host's element id, units are the host's internal units, and the script is not a record of the building. KIR emits code for the first of them; its argument is that emitted code should be derived from a typed program rather than written.

**Exchange and versioning of state.** IFC [8, 9] and IfcOpenShell [10] describe building state in a host-neutral schema with stable global identifiers; round-trip studies document geometric distortion and loss of authoring semantics [11]. IFCdiff [12] and graph-based version control [13, 14] address differencing and asynchronous collaboration at the level of exchanged state. KIR's journal of observed state is closest to this line; its states are multisets of operations in a registry that also compiles, so a delta is executable against the host rather than only descriptive.

**Procedural and object-model authoring.** Grasshopper [15] and the parametric-design tradition [16] treat the building as a dataflow program, and Davis studies why such models become brittle under later edits [17]. Speckle [18, 19], BHoM [20, 21], Hypar Elements [22] and COMPAS [23] supply typed or graph-shaped vocabularies for buildings in code. Among the systems reviewed we found none that places a compiler with transaction, refusal and witness semantics between the description and a stateful host; they are candidate producers for KIR rather than alternatives (Supplement Table S3 compares where each approach puts the abstraction boundary).

**Language models and BIM.** Text2BIM drives a BIM tool through a multi-agent framework that generates API code and refines the model with a rule-based checker [24]; BIMgent operates the authoring tool's graphical interface and reports a 32% task success rate against a 0% baseline [25]; BIM-GPT addresses retrieval rather than authoring [26]. The Model Context Protocol standardises discovery and invocation of tools with JSON-schema arguments [27, 28], and a reference architecture for BIM servers adds an adapter contract with versioned files and diffs on IfcOpenShell [29]; neither specifies refusals with candi-dates, per-field origins or unknown transaction outcomes — the gap this paper studies.

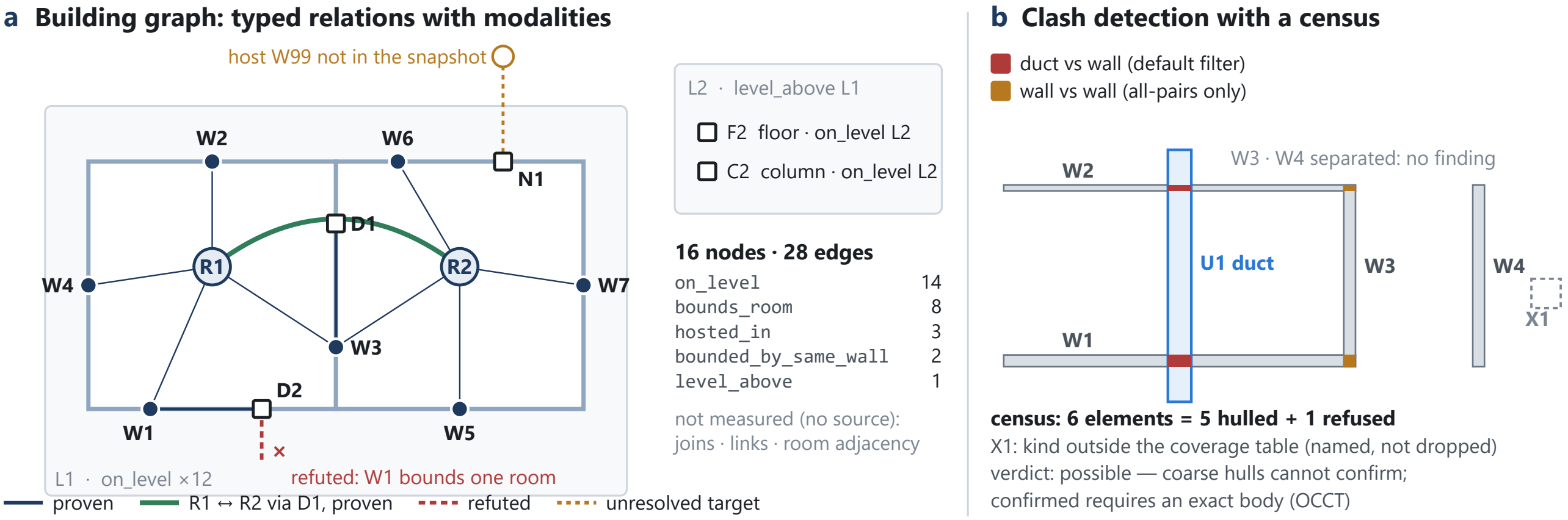

**Figure 8.** Coordination on the representation: KIR 0.8.1's graph and clash code run on a small synthetic document built for this figure (the project's own scenes: Supplement Section S4). (a) 16 nodes and 28 typed edges over a schematic plan, the 14 `on_level` edges drawn as level bands: D1 in the shared wall W3 gives a proven room-to-room relation, D2 in the exterior wall W1 the same relation refuted by a named rule, and N1's host outside the extraction an unresolved edge. (b) Six elements: five hulls and one named refusal balance the census; the duct meets both walls under the default filter, the wall junctions only under the all-pairs filter, and the separated walls produce no finding.

**CAD intermediate representations.** CADIR [3] is an executable, editable IR for agentic mechanical CAD with export to FreeCAD, SolidWorks and Fusion 360; ArtisanCAD [4] uses the name CAD-IR for a component of an industrial CAD agent. Both are concurrent with this work. KIR differs in its target — hosted, levelled elements in a persistent document with host defaults, native transactions and partial readback — and in the vocabulary that target requires. DeepCAD [30], SketchGraphs [31] and the Fusion 360 Gallery [32] treat construction sequences as learning data, and CADParser recovers sequences from geometry [33]; see [34] for a survey.

**Compilers and certified code.** Deriving several surfaces from one declaration and making a class of errors inexpressible by types is ordinary compiler practice: LLVM [35], MLIR [36], Halide's separation of what from how [37] and typed assembly language [38]. Proof-carrying code [39] ships a proof with the code so that the consumer can check it; KIR's certificate and witnesses are a weaker, empirical cousin — evidence that a check exists and can fail, and host readback after execution, rather than a proof checked before it.

**Infrastructure as code, reconciliation and build systems.** The closest systems analogy is infrastructure as code: Terraform separates a reviewable plan from its application and records state in order to detect drift [40], and cluster managers reconcile declared and observed state continuously [41]. KIR's plan digest bound to a snapshot, its document probe and its separation of authored from observed history follow the same logic. The differences are in the target: a BIM host is a licensed, in-process document with no declarative API, so KIR lowers to imperative host code and must gather evidence per element. Treating the host as a build target also places KIR in the design space of build systems [42], with the unusual property that the target can report an outcome as unknown.

**Transactions, concurrency and provenance.** The state `UNCONFIRMED` is the in-doubt state of distributed transaction processing [43, 44], made a first-class value with its own retry rule. The store's head check is the value-comparison half of optimistic concurrency control [45]; stamps and the operation journal follow Helland's idempotence discipline [46]; content addressing follows Merkle and IPFS [47, 48]; the merge follows the software-merging tradition [49]. `FieldOrigin` is a coarse, per-field form of where-provenance [50]. The judge descends from automated rule checking [51], with the difference that it names the rules it could not evaluate.

**Upstream producers.** Floor-plan recognition [52–54], reconstruction from scanned plans [55], scan-to-BIM [56, 57] and constrained reconstruction from images and point clouds [58, 59] produce geometry and semantics that must eventually be committed to a host. Floor-SP [60] frames reconstruction from RGB-D scans as inverse CAD, optimising a floorplan graph room by room with terms that encourage adjacent rooms to share corners and walls. Such systems recover structure from observations; KIR begins downstream, where that structure becomes a program to be refused, recorded, witnessed and versioned before it becomes a model.

## 10 Discussion and Limitations

**Why not direct API code?** The strongest form of the objection is serious: current models write idiomatic .NET, the API is documented, a compile-and-repair loop converges, and any intermediate layer is a second implementation that can be wrong, lags the API and confines the agent to what its registry contains — 83 operations on the snapshot. Our answer is not that a registry can replace the API today. It is that direct code has no native place to hold the information of Section 2: a compile-and-repair loop can fix code that fails to compile, but it does not reveal a selector that happened to choose an object, a default whose origin vanished, a check that never ran, or a transaction whose response was lost after commit — and a repair loop cannot repair what compiles. Two further properties matter to a bureau: validation against the host is a licensed, serial round trip, and two agents' emitted scripts cannot be merged while two agents' programs can. Whether these mechanisms yield higher end-to-end success under equal budgets is precisely RQ7.

**Why not IFC?** IFC represents the state of a building; KIR represents operations against a host that holds state, with what must be true before and after each and what was decided on the way. IFC has no snapshot against which a name is ambiguous, no transaction, no witness and no unknown outcome, and an import need not preserve native authoring constructs, whereas KIR's emitters create native elements through the host's own API.

**Is this defensive programming with a vocabulary?** Every mechanism of Section 2 is good engineering practice. The difference is placement: when the vocabulary lives in the representation and the outcome type rather than in a log, downstream consumers can act on it — a repair loop reads a refusal's candidates, a journal stores an unknown outcome and refuses a replay, a judge reads origins to know which values were authored. The discipline also has a measured cost: test files are 47.1% of the package's 556,060 lines.

**What KIR does not prove.** A well-typed, committed and fully witnessed program can still describe a bad building: the corpus contains a three-millimetre wall that passes the unit discipline, and the judge has only twenty rules and often lacks inputs to evaluate them. KIR's contribution is the separation of validity, execution evidence and design correctness; it does not collapse the third into the first two. Table 3 lists the principal limitations and what would retire each; Supplement Section S8 lists more.

## 11 Conclusion

KIR asks what must be carried between an autonomous author and a stateful, partially observable BIM host. The answer is more than geometry or API calls: ambiguity, provenance, named absence, unknown transaction state, reverse-path loss, stale premises and authored history, in forms another agent — and ultimately a human signer — can inspect and act on. On the studied snapshot the implementation demonstrates a registry-derived typed language, host-free refusal and rehearsal, a static guard against vacuous witnesses, an outcome space specified by seven invariants, content-addressed revisions under compare-and-swap, and a reverse path that cannot silently drop an element. The offline results are reproducible from the published scripts; the native Revit and Roslyn results are project-reported. The central missing test is equally explicit: a controlled comparison with agents that write host code directly, across stateful edits and forced recovery. If that experiment and a second backend succeed, KIR will move from a systems argument for an evidence-carrying BIM representation to a demonstration that such a representation improves agent-authored modelling.

**Outlook.** Four directions follow, each measurable in the paper's own vocabulary. An IFC writer for the expressible subset, reporting the contracts that have no IFC analogue as named absences, would be the second backend that tests independence from Revit. Perception front ends — drawing recognition and scan-to-BIM — can emit programs under the census invariant, so that drawings and scans of one building merge as typed proposals against one store. Checked contracts between disciplines, a slot one agent leaves and another must fill within a bound, would turn the bureau's declared scopes into refusals. And building the graph and the clash hulls of Figure 8 from a grounded program rather than from extracted state would let an agent learn what a proposed wall bounds or hits before anything is materialised. Supplement Section S4 lists further directions.

**Code and data availability.** KIR is open source (Apache-2.0) at github.com/5vbkgsghhh-hash/kir and on PyPI as `kir-building`. Scripts and offline outputs of the experiments are in its `paper/` directory (Supplement Section S6); RQ6's native-run corpora are not public.

**Use of generative AI.** The research and the source manuscript are the author's. Language-model assistants helped condense and edit the text, draft some passages, typeset the paper, write the figure scripts and re-run the checks in Supplement Table S9; the author reviewed all of it and takes full responsibility.

## References

[1] Autodesk. *Revit API Developer's Guide and SDKs*. Autodesk Platform Services, 2024. https://aps.autodesk.com/developer/overview/revit-api

[2] Autodesk. Introducing the Revit Public MCP Server: a trusted foundation for AI-powered workflows. *AEC Tech Drop* (blog), June 2026.

[3] Y. Liu, J. Ni, Y. Chen, J. Huang, R. Tong, M. Tang, and P. Du. CADIR: a cross-backend editable intermediate representation for agentic CAD generation. arXiv:2608.00891, 2026.

[4] Y. Xu, Q. Wu, X. Li, Y. Bin, Q. Yao, J. Gu, G. Wang, W. Lv, H. Yang, W. Luo, J. Xiang, Y. Chen, and S. Chen. ArtisanCAD: an industrial-level CAD agent with expert-grounded knowledge distillation. arXiv:2607.05750, 2026.

[5] Architecture and Building Systems group. RevitPythonShell: an IronPython console for the Revit API. https://github.com/architecture-building-systems/revitpythonshell, 2026.

[6] pyRevitLabs. pyRevit: rapid application development for Autodesk Revit. https://github.com/pyrevitlabs/pyRevit, 2026.

[7] Autodesk. Dynamo: visual programming for Revit. https://dynamobim.org, 2026.

[8] ISO 16739-1:2024. *Industry Foundation Classes (IFC) for data sharing in the construction and facility management industries — Part 1: Data schema*. International Organization for Standardization, 2024.

[9] buildingSMART International. Industry Foundation Classes (IFC). https://technical.buildingsmart.org/standards/ifc/, 2026.

[10] IfcOpenShell contributors. IfcOpenShell: open-source IFC toolkit and geometry engine. https://ifcopenshell.org, 2026.

[11] T. Pazlar and Ž. Turk. Interoperability in practice: geometric data exchange using the IFC standard. *Journal of Information Technology in Construction (ITcon)*, 13:362–380, 2008.

[12] X. Shi, Y.-S. Liu, G. Gao, M. Gu, and H. Li. IFCdiff: a content-based automatic comparison approach for IFC files. *Automation in Construction*, 86:53–68, 2018.

[13] S. Esser, S. Vilgertshofer, and A. Borrmann. Graph-based version control for asynchronous BIM level 3 collaboration. In *Proc. EG-ICE Workshop on Intelligent Computing in Engineering*, pages 98–107, 2021.

[14] S. Esser, S. Vilgertshofer, and A. Borrmann. Graph-based version control for asynchronous BIM collaboration. *Advanced Engineering Informatics*, 53:101664, 2022.

[15] Robert McNeel & Associates. Grasshopper: algorithmic modeling for Rhino. https://developer.rhino3d.com, 2026.

[16] R. Woodbury. *Elements of Parametric Design*. Routledge, 2010.

[17] D. Davis. *Modelled on Software Engineering: Flexible Parametric Models in the Practice of Architecture*. PhD thesis, RMIT University, 2013.

[18] Speckle Systems. Speckle: the open-source data platform for AEC. https://speckle.systems, 2026.

[19] P. Poinet, D. Stefanescu, N. Tsakiridis, X. de Boisse, and E. Papadonikolaki. Supporting collaborative design and project management for AEC using Speckle's interactive data flow diagram. In *Design Computation Input/Output (DC I/O)*, London, 2020.

[20] BuroHappold Engineering and the BHoM community. BHoM: the Buildings and Habitats object Model. https://github.com/BHoM/BHoM, 2026.

[21] D. Elshani, A. Lombardi, A. Fisher, D. Hernández, S. Staab, and T. Wortmann. Knowledge graphs for multidisciplinary co-design: introducing RDF to BHoM. In *Proc. LDAC Workshop*, CEUR Workshop Proceedings 3213, pages 32–42, 2022.

[22] Hypar. Hypar Elements: the smallest useful BIM. https://github.com/hypar-io/Elements, 2026.

[23] COMPAS contributors. COMPAS: a computational framework for collaboration and research in architecture, engineering, fabrication and construction. https://compas.dev, 2026.

**Table 3.** Principal limitations of the snapshot and the evidence that would retire each.

| Limitation | Consequence | Required next evidence |
|---|---|---|
| RQ7 not executed | no controlled claim that KIR beats directly generated C# | same-model, same-budget live comparison on second and third edits, recovery and preservation |
| One emission backend (Revit) | backend independence is a property of form only | a second target, preferably IFC for the expressible subset |
| Native corpora not shipped | readers cannot replay receipts and witnesses | release documents, receipts, witnesses and extracts of the native runs |
| Units typed, not plausibility-checked | metre-intended values compile as millimetres | contextual plausibility diagnostics without silent coercion |
| Pinned `element_id` checked at runtime | one reference-error class is not rejected offline | snapshot identity sufficient to validate pinned selectors before emission |
| Thin witness coverage | 16 operations witnessed on one axis; queries on none | wider independent readback; keep named absences where proof is unavailable |
| Hand-written emitter slot access | registry changes are not mechanically propagated to emitters | declarative backend descriptions with generated slot binding |
| Project-authored evaluation | tests are extensive but not independent | third-party reproduction and a user study |

[24] C. Du, S. Esser, S. Nousias, and A. Borrmann. Text2BIM: generating building models using a large language model-based multi-agent framework. *Journal of Computing in Civil Engineering*, 40(2), 2026. arXiv:2408.08054.

[25] Z. Deng, C. Du, S. Nousias, and A. Borrmann. BIMgent: towards autonomous building modeling via computer-use agents. arXiv:2506.07217, 2025.

[26] J. Zheng and M. Fischer. BIM-GPT: a prompt-based virtual assistant framework for BIM information retrieval. arXiv:2304.09333, 2023.

[27] Model Context Protocol contributors. Model Context Protocol specification. https://modelcontextprotocol.io, 2026.

[28] Anthropic. Introducing the Model Context Protocol. https://www.anthropic.com/news/model-context-protocol, November 2024.

[29] T. Heimig-Elschner, C. Du, A. Scheuvens, A. Borrmann, and J. Beetz. A modular reference architecture for MCP-servers enabling agentic BIM interaction. In *Proc. GNI Symposium on Artificial Intelligence for the Built World*, Technical University of Munich, 2026. arXiv:2601.00809.

[30] R. Wu, C. Xiao, and C. Zheng. DeepCAD: a deep generative network for computer-aided design models. In *Proc. IEEE/CVF International Conference on Computer Vision (ICCV)*, 2021.

[31] A. Seff, Y. Ovadia, W. Zhou, and R. P. Adams. SketchGraphs: a large-scale dataset for modeling relational geometry in computer-aided design. arXiv:2007.08506, 2020.

[32] K. D. D. Willis, Y. Pu, J. Luo, H. Chu, T. Du, J. G. Lambourne, A. Solar-Lezama, and W. Matusik. Fusion 360 Gallery: a dataset and environment for programmatic CAD construction from human design sequences. arXiv:2010.02392, 2020.

[33] S. Zhou, T. Tang, and B. Zhou. CADParser: a learning approach of sequence modeling for B-Rep CAD. In *Proc. International Joint Conference on Artificial Intelligence (IJCAI)*, 2023.

[34] L. Zhang, B. Le, N. Akhtar, S.-K. Lam, and T. Ngo. Large language models for computer-aided design: a survey. arXiv:2505.08137, 2025.

[35] C. Lattner and V. Adve. LLVM: a compilation framework for lifelong program analysis & transformation. In *Proc. International Symposium on Code Generation and Optimization (CGO)*, pages 75–86, 2004.

[36] C. Lattner, M. Amini, U. Bondhugula, A. Cohen, A. Davis, J. Pienaar, R. Riddle, T. Shpeisman, N. Vasilache, and O. Zinenko. MLIR: scaling compiler infrastructure for domain specific computation. In *Proc. CGO*, 2021.

[37] J. Ragan-Kelley, C. Barnes, A. Adams, S. Paris, F. Durand, and S. Amarasinghe. Halide: a language and compiler for optimizing parallelism, locality, and recomputation in image processing pipelines. In *Proc. ACM SIGPLAN PLDI*, pages 519–530, 2013.

[38] G. Morrisett, D. Walker, K. Crary, and N. Glew. From System F to typed assembly language. *ACM Transactions on Programming Languages and Systems*, 21(3):527–568, 1999.

[39] G. C. Necula. Proof-carrying code. In *Proc. 24th ACM SIGPLAN-SIGACT Symposium on Principles of Programming Languages (POPL)*, pages 106–119, 1997.

[40] HashiCorp. Terraform documentation: plans, applies and state. https://developer.hashicorp.com/terraform/docs, 2026.

[41] B. Burns, B. Grant, D. Oppenheimer, E. Brewer, and J. Wilkes. Borg, Omega, and Kubernetes. *ACM Queue*, 14(1):70–93, 2016.

[42] A. Mokhov, N. Mitchell, and S. Peyton Jones. Build systems à la carte. *Proceedings of the ACM on Programming Languages*, 2(ICFP):79:1–79:29, 2018.

[43] J. Gray and A. Reuter. *Transaction Processing: Concepts and Techniques*. Morgan Kaufmann, 1993.

[44] P. A. Bernstein, V. Hadzilacos, and N. Goodman. *Concurrency Control and Recovery in Database Systems*. Addison-Wesley, 1987.

[45] H. T. Kung and J. T. Robinson. On optimistic methods for concurrency control. *ACM Transactions on Database Systems*, 6(2):213–226, 1981.

[46] P. Helland. Idempotence is not a medical condition. *ACM Queue*, 10(4):30–46, 2012.

[47] R. C. Merkle. A digital signature based on a conventional encryption function. In *Advances in Cryptology — CRYPTO '87*, LNCS 293, pages 369–378, 1988.

[48] J. Benet. IPFS — content addressed, versioned, P2P file system. arXiv:1407.3561, 2014.

[49] T. Mens. A state-of-the-art survey on software merging. *IEEE Transactions on Software Engineering*, 28(5):449–462, 2002.

[50] P. Buneman, S. Khanna, and W.-C. Tan. Why and where: a characterization of data provenance. In *Proc. International Conference on Database Theory (ICDT)*, LNCS 1973, pages 316–330, 2001.

[51] C. M. Eastman, J. Lee, Y. Jeong, and J. Lee. Automatic rule-based checking of building designs. *Automation in Construction*, 18:1011–1033, 2009.

[52] C. Liu, J. Wu, P. Kohli, and Y. Furukawa. Raster-to-vector: revisiting floorplan transformation. In *Proc. ICCV*, pages 2214–2222, 2017.

[53] Z. Zeng, X. Li, Y. K. Yu, and C.-W. Fu. Deep floor plan recognition using a multi-task network with room-boundary-guided attention. In *Proc. ICCV*, pages 9095–9103, 2019.

[54] A. Kalervo, J. Ylioinas, M. Häikiö, A. Karhu, and J. Kannala. CubiCasa5K: a dataset and an improved multi-task model for floorplan image analysis. In *Proc. Scandinavian Conference on Image Analysis (SCIA)*, LNCS 11482, pages 28–40, 2019.

[55] L. Gimenez, S. Robert, F. Suard, and K. Zreik. Automatic reconstruction of 3D building models from scanned 2D floor plans. *Automation in Construction*, 63:48–56, 2016.

[56] P. Tang, D. Huber, B. Akinci, R. Lipman, and A. Lytle. Automatic reconstruction of as-built building information models from laser-scanned point clouds: a review of related techniques. *Automation in Construction*, 19(7):829–843, 2010.

[57] Y. Wu and F. Xue. FloorPP-Net: reconstructing floor plans using point pillars for scan-to-BIM. arXiv:2106.10635, 2021.

[58] F. Xue, W. Lu, and K. Chen. Automatic generation of semantically rich as-built building information models using 2D images: a derivative-free optimization approach. *Computer-Aided Civil and Infrastructure Engineering*, 33(11):926–942, 2018.

[59] F. Xue, W. Lu, C. J. Webster, and K. Chen. A derivative-free optimization-based approach for detecting architectural symmetries from 3D point clouds. *ISPRS Journal of Photogrammetry and Remote Sensing*, 148:32–40, 2019.

[60] J. Chen, C. Liu, J. Wu, and Y. Furukawa. Floor-SP: inverse CAD for floorplans by sequential room-wise shortest path. In *Proc. ICCV*, pages 2661–2670, 2019.

SUPPLEMENTARY MATERIAL

# A Building as a Repository: KIR

*Implementation map, full results, protocols, reproduction and a cross-environment re-derivation*

Dmitry Kuklev · Independent Researcher · dima.kuklev9797@gmail.com · code: github.com/5vbkgsghhh-hash/kir

All snapshot-specific statements refer to release KIR 0.8.1 (2026-09-08) unless stated otherwise; the release is archived on PyPI as `kir-building` 0.8.1, while the repository continues to change. This document complements the main paper and does not repeat its narrative; section, figure and table numbers prefixed with S belong to this supplement.

## S1 Scope and conventions

Evidence levels follow the main paper: **IMPL** (source-level implementation), **TEST-OFF** (covered by an offline test with fixtures), **EMIT** (present in emitted C#), **ROSLYN** (compiled against Revit API assemblies by the project's compile service; project-reported), **LIVE** (executed in native Revit; project-reported from operational records) and **FUTURE** (planned). File and function references are to that release. Where a number changed between the previous release measured with the same scripts (KIR 0.6.0, 2026-09-04) and the studied one, the earlier value is given in brackets.

## S2 Implementation map and design requirements

KIR is a Python 3.12 package under the Apache-2.0 licence, distributed on PyPI as `kir-building`. On the snapshot it has 376 production modules and 911 test files (294,081 and 261,979 lines; 301 and 694 files on 0.6.0), four direct core dependencies (shapely, networkx, pydantic and httpx, each bounded below the next major version) and an optional OCCT geometry kernel used by the exact clash phase, the geometry loop and refinement deviation; every code path that needs the kernel reports `kernel_unavailable` by name when it is absent. Table S1 maps the paper's mechanisms to modules. Table S2 lists twelve conventional properties of a good program representation — distinct from the seven bureau requirements of main-paper Section 2, which subsume them — with the mechanism, evidence and remaining gap for each.

**Table S1.** Implementation map (paths relative to the package root `kir/`).

| Layer | Primary modules | Mechanisms described in the paper |
|---|---|---|
| Typed carrier | `registry_base.py`, `spec.py`, `ops_*.py`, `sdk.py`, `dsl.py`, `macros.py`, `schema_gen.py`, `op_contract.py` | 83 contracts; 39 parameter kinds; 8 reference kinds; generated SDK, DSL and schema; contract digest `kir-op-contract/2` |
| Planning and grounding | `compiler.py`, `midend.py`, `ground.py`, `relate.py`, `chunking.py` | envelope and type checks; backward-only references; `FieldOrigin`; grounding with candidates; plan, contract and grounding digests |
| Rehearsal and certificate | `rehearsal.py`, `translation_cert.py`, `emit_model.py` | obligation table (377 clauses); named absences (38); reachability certificate; witness objects that require a verdict |
| Emission | `authoring.py`, `*_emit.py`, `emit_core.py`, `emit_transaction_unit.py`, `document_guard.py` | three dispatch tables; per-version branches; unit conversion in emitted code; document probe; stamps |
| Execution semantics | `outcome.py`, `contracts.py`, `bridge_result.py`, `acceptance*.py`, `a5_*.py`, `idempotence*.py` | outcome space; typed receipts; `X004` vs `W004`; recovery journal; fail-closed rebuild gate |
| Authored persistence | `project.py`, `project_store.py`, `project_merge.py`, `project_diff.py`, `project_refinement.py` | content-addressed revisions; compare-and-swap; typed proposals; three-way merge; diff by address; lineage |
| Observed state | `decompile/` (census, lift, fold, journal, merge3, capture) | census invariant; typed atoms; identity-free journal; delta rebuild; offline capture editing |
| Coordination | `decompile/building_graph.py`, `clash/`, `refine/`, `checker/`, `design_check.py`, `bureau/`, `mcp/` | typed relations with modalities; hull and exact clash; judge with thresholds; bureau of workers; MCP door |

**Table S2.** Twelve conventional representation properties: mechanism, evidence level and remaining gap on the snapshot.

| Property | Mechanism | Evidence | Remaining gap |
|---|---|---|---|
| R1 explicit identity | `id` field unique per program (`KIR-P006`); output ids independent of position and content in the store | IMPL, TEST-OFF | observed-state identity derives from host element ids |
| R2 typed references | eight reference kinds; slot-level accepted kinds; `KIR-L004` | IMPL, TEST-OFF | View, Sheet, Room and Grid kinds promised by the written specification are absent |
| R3 deterministic semantics | canonical plan digest; `FieldOrigin`; grounding report | IMPL; byte-identical re-emission | host-side defaults lie outside the record |
| R4 unit discipline | millimetre and degree kinds; no feet kind; conversion in emitted code | IMPL | no plausibility check (3.0 mm accepted) |
| R5 explicit effects | effect kind per operation; write flags; destructive gate (`KIR-D001`) | IMPL | effects scheduler implemented and tested but not wired |
| R6 dependency ordering | backward-reference invariant, one forward scan (`KIR-L003`) | IMPL, TEST-OFF | no graph object; forward references illegal |
| R7 backend independence | registry names no host API member; version is a compiler input | IMPL (form only) | only Revit emitters exist |
| R8 inspectability | JSON program; planned program; grounding and rehearsal reports; offline preview | IMPL | the plan is a list, not a graph |
| R9 persistence across edits | content-addressed revisions with compare-and-swap; proposals; journal with delta reuse | IMPL, TEST-OFF | durability tests simulate process crash, not power loss |
| R10 refusal under ambiguity | `KIR-G101`–`G104` with candidates | IMPL, TEST-OFF | `element_id` passed to a runtime guard |
| R11 execution provenance | stamp `kir:<hash>:<op_id>`; typed receipt; connector operation journal | IMPL + EMIT | stamp written after creation (finding F-244 open) |
| R12 completion ≠ verified intent | three-axis outcome; `X004` vs `W004`; `PROGRAM` vs `PARSE` verdict | IMPL, TEST-OFF | live path not reproducible from the snapshot |

Table S3 places KIR among the approaches reviewed in main-paper Section 9 by where each puts the abstraction boundary: what the author writes, where identity lives, who fixes units and the API version, what “done” means and what persists.

**Table S3.** Where the abstraction boundary sits in four approaches. KIR does not take responsibility for geometry kernels, visual programming, host-neutral exchange (IFC remains the format) or design correctness.

| Approach | Author writes | Identity lives in | Units and API version | “Done” means | What persists |
|---|---|---|---|---|---|
| Direct API code (C#, pyRevit, Dynamo) | host API calls | host element ids in variables | the author, per line | the call returned without exception | the script; the document |
| IFC-first (IFC, IfcOpenShell) | a description of state | a global id per entity | fixed by the schema; no API version | the file validates; import may be lossy | the file |
| Object models (Grasshopper, Hypar, BHoM, Speckle) | a dataflow or typed object graph | the model's own object ids | converters per tool | objects converted | the graph; object commits |
| KIR | a Python script producing a typed program | `op.id` in the program; host id only after execution | the compiler (mm only; version as input) | a typed outcome on three axes; unknown is a state | revisions of the program; a journal of observed state |

## S3 Full results

Table S4 lists every experiment with its denominator. The paragraphs below give details that the main paper summarises.

**Table S4.** Research questions, experiments, denominators and results (0.6.0 values in brackets where changed).

| RQ | Experiment | Denominator | Result | Evidence |
|---|---|---|---|---|
| RQ1 | signatures, schema, docs vs registry | 83 operations | 0 / 0 / 0 mismatches | IMPL |

| RQ | Experiment | Denominator | Result | Evidence |
|---|---|---|---|---|
| RQ1 | emitter slot access | 77 writing emitters | 297 literal vs 0 registry-driven | source scan |
| RQ1 | one program per operation × 6 versions | 498 attempts (83 × 6) | 482 emitted; 16 typed version refusals over 7 operations; 76 [75] operations on all six | TEST-OFF |
| RQ2 | 41 stress variants + valid baseline | 42 programs | 29 refused; 27/29 name operation and field; 0 exceptions; 3 silent | TEST-OFF |
| RQ3 | 5 programs × 6 versions: room, room with door, tower (3 programs) | 30 emissions | 30 ok, 0 diagnostics; branches at 2022 (tower) and 2024 | TEST-OFF |
| RQ3 | compactness | room; tower | 612 vs 34,420 (56×); 11,263 vs 3,709,235 (329× [301×]) | IMPL |
| RQ3 | determinism | 1 program × 3 runs | three equal SHA-256 digests | IMPL |
| RQ4 | project-store sequence | 3 revisions, 3 proposals | stale commit refused; R2 accepted; wrong head refused; stale base conflict | TEST-OFF |
| RQ4 | journal of observed state | 5–6 leaves; 2 merges | 4/5 and 4/6 reused; 1 conflict, 1 clean | TEST-OFF |
| RQ4 | test files | 233 + 186 tests | all pass (31 kernel-gated skips) | TEST-OFF |
| RQ5a | omitted vs explicit default; tolerance mutation | 1 wall; 1 operation | plan digest differs; contract digest changes and restores | IMPL |
| RQ5b | obligation table; rehearsal per operation | 77 writes; 83 programs | 377 [374]; 38 named (27/8/3) [36]; 3/16/31/27/3 operations on 0–4 axes | IMPL |
| RQ5c | certificate over golden corpus | 73 programs × 3 versions | 63/68/69 certified; 3 uncertifiable; 7/2/1 gated; 0 vacuous | TEST-OFF |
| RQ5d | enumerate outcome space | 5 × 4 × 5 = 100 | 31 admitted; retry 14/13/4 | IMPL |
| RQ5e | census laws; 92-category document | 130 tests; 92 elements | all pass (2 skips); 8 + 84 = 92; 5/5 laws tested | TEST-OFF |
| RQ5f | judge, instruments, agreements | 20 rules; 13 instruments; 10 agreements | 7 evaluated / 13 not; 7/3/3/0; 9 hold, 1 finding | IMPL, TEST-OFF |
| RQ5g | graph, clash, stages, thresholds, bureau, capture | synthetic documents; test suites | see §S4 | IMPL, TEST-OFF |
| RQ6 | project operational runs | 28 + 21 operations; 1,242 compilations | 26/28; 17/21 with 4 classified; 1,242/1,242 | LIVE, ROSLYN |
| RQ7 | direct-code baseline | — | protocol specified (§S5); not run | FUTURE |

### S3.1 RQ1: consistency and the emitter seam

The operation added since 0.6.0 is the query `query_element_state`. Emitter dispatch is by three hand-maintained tables (72 + 5 + 6 = 83); Table S5 counts string-literal slot references per file, a lower bound obtained by a regular expression over each emitter's source. Per-version coverage: of 498 attempts (83 × 6), 482 emit (76 operations on all six versions, 3 on five, 3 on three, 1 on two) and 16 are typed `KIR-E003` refusals over seven operations, for example ceilings before Revit 2022, free-standing loads from 2024 and multistorey stairs before 2025.

**Table S5.** Hand-written slot references in the 77 writing emitters.

| File | Operations | Literal slot accesses | Registry-driven |
|---|---|---|---|
| `authoring.py` | 35 | 160 | 0 |
| `struct_emit.py` | 6 | 34 | 0 |
| `solid_emit.py` | 4 | 21 | 0 |
| `analysis_emit.py` | 4 | 13 | 0 |
| `datum_emit.py` | 3 | 13 | 0 |
| `arch_emit.py` | 2 | 11 | 0 |
| `boolean_emit.py` | 1 | 11 | 0 |
| `mep_emit.py` | 5 | 7 | 0 |
| five single-operation files | 5 | 6 | 0 |
| `surface_emit.py`, `shape_emit.py` | 2 | 8 | 0 |

| File | Operations | Literal slot accesses | Registry-driven |
|---|---|---|---|
| room_emit.py, mass_emit.py, site_emit.py | 6 | 9 | 0 |
| sweep_emit.py, adaptive_emit.py, opening_emit.py | 4 | 4 | 0 |
| **Total** | **77** | **297** | **0** |

### S3.2 RQ2: corpus design and outcomes

The baseline program contains a level, four walls, a door and a window hosted on two of them, and a room, compiled against a snapshot with one door symbol (id 700), one window symbol (id 800) and one level (id 30, elevation 0). The 41 variants fall into eight categories; Table S6 is the table view of main-paper Figure 6a. Results were identical on both snapshots. The DSL front end, probed on nine further cases, refuses structural defects at authoring time (duplicate id, missing or unknown argument, unknown operation) and passes semantic ones to the compiler, which refuses them.

**Table S6.** RQ2: categories, outcomes and diagnostic codes observed (42 programs).

| Category | n | Refused | Accepted | Codes observed |
|---|---|---|---|---|
| Baseline | 1 | 0 | 1 (intended) | — |
| Legal variations | 7 | 1 | 6 (intended) | `P003` (closed envelope) |
| Reference errors | 9 | 8 | 1 silent: `element_id` absent from snapshot | `L003`, `L004`, `P006`, `G102` |
| Parameter types | 9 | 9 | 0 | `T001`, `T002`, `P002`–`P004` |
| Geometry | 6 | 5 | 1 benign | `T001`, `T002` |
| Grounding | 3 | 3 | 0 | `G101`, `G103`, `G104` |
| Host / dependency | 3 | 2 | 1 unreachable placeholder | `L003`, `L004` |
| Units | 2 | 0 | 2 silent: metre-intended lengths | — |
| Version capability | 2 | 1 | 1 control | `E003` |
| **Total** | **42** | **29** | **13 (10 intended, 3 silent)** | **13 distinct codes** |

### S3.3 RQ5: the evidence mechanisms in detail

*Named absences.* The 38 named clauses comprise 27 exemptions (pre-flight refusals, host-derived values, emission-order rules), 8 absences proper (rooms bounded by a moved element; the geometry of both operands after a join, which shifts solids by half a thickness; a building pad's boundary, whose arc representation is unmeasured; a collateral deletion that is reported but not bounded; a curtain panel count; neighbour circles after a type change or a parameter write) and 3 caveats (a bounding box untrustworthy when a ring carries a spline whose tangents the host does not document; a riser count legitimately short before a landing completes the flight). A live count in the serving layer gives the same picture from the other side: of 77 writing operations, 14 declare no geometry obligation, 24 no topology obligation and 31 no semantic obligation, and that list travels with every success receipt. Listing S1 reproduces the rehearsal report of the demonstration house.

**Listing S1.** Rehearsal report of the demonstration house (a level, a floor, four walls, a room), computed without any host; reproduced from the source manuscript.

```
REHEARSAL · operations declared 7 · elements after group expansion 7 · flights to Revit 0

VALUES THE HOST WILL OVERWRITE (authority = DERIVED_BY_REVIT)
   none

AUTHORITY PASSES SILENTLY TO ANOTHER MECHANISM (field omitted)
   ×4 create_wall: top_level omitted
        will not be checked: top constraint == resolved top_level when given (topology)
        will not be checked: built wall spans at least base..top elevation (geometry)
        decided instead by: the top is a NUMBER (height_mm) and no longer follows the level;
        a later shift of the level does not move the wall
   ×1 create_room: upper_offset_mm omitted
        will not be checked: upper offset param == upper_offset_mm when given (geometry)
        decided instead by: the document type's default (usually 8 ft = 2 438 mm),
        below the habitability floor HAB022 (min ceiling height 2 500 mm) almost always

WILL NOT BE CHECKED -- NAMED BY THE PROJECT (a decision with a reason)
   create_room: placed after -- emit-order rule (Regenerate before rooms), not a post-commit witness

WRITES, BUT WATCHED ON EXACTLY ONE AXIS
   none

optional capabilities not used -- 18 obligations over 6 fields (normal, not a finding)

WILL BE CHECKED
   ×2 create_floor · geometry   ×1 create_floor · semantic   ×1 create_floor · topology
   ×1 create_level · geometry   ×1 create_level · identity
   ×1 create_room · geometry    ×1 create_room · identity    ×1 create_room · semantic
   ×1 create_room · topology
   ×4 create_wall · geometry    ×4 create_wall · parameter   ×4 create_wall · topology
```

*Judge.* A rule falls silent for one of four named reasons: a precondition input the model cannot supply; a subject set that is non-empty in principle but wholly withheld (a case the engine's header records as previously mis-counted); a stage profile that suspends the rule; or a degenerate model, which forces a blocking verdict rather than a pass. On the README example, the verdict refuses a living room with no exterior window and lists thirteen of twenty rules as not evaluated because the program contains no stairs and no ceiling height. *Instruments.* The parameter-boundary audit counts 757 boundaries (83 registry, 85 tolerances, 343 named constants, 24 references, 222 literals). *Agreements.* Ten self-consistency checks between code paths, each with a mandatory demonstration that it can fail; nine hold and one reports a live finding. Table S7 lists the five conservation laws.

**Table S7.** The five conservation laws: enforcing code and tests on the snapshot.

| Law | Enforcing code | Test | Result |
|---|---|---|---|
| Census: lifted + atoms + unread = document count | `census.reconcile_census`; run-stopping error in the pipeline | `test_census_law.py` | 28 passed |
| Receipt: every cut leaves a row or a failure | `side_contract.reconcile_side_stage`; rows + failures in five extractors | `test_receipt_law.py` | 27 passed |
| Witness axis: sign only what was read | `translation_cert.py`; per-operation readers | `test_witness_axis_honesty.py` | 2 + 8 subtests |
| Contamination: a partial read marks what derives from it | `is_partial_read`, propagated in extraction and pipeline | `test_partial_read_contagion.py` | 5 passed |
| Neutrality: no device ids or foreign paths in the compiler | source scan inside the test | `test_supply_neutrality.py` | 8 passed, 2 skipped |

### S3.4 RQ6: live records and artefacts

The Revit 2023 matrix of 2026-07-27 (artefact `live_op_matrix_20260727.json`) ran against a production model; the broken operation was `create_dimension`, later repaired, and `create_beam` had no substrate in that model. The compile-service records are `A5_compile_gate_full.json` and `A6_gate_final.json`. The Revit 2026 matrix of 2026-09-03 (`live_op_matrix_gap.json`) ran through the MCP door with per-operation geometry and topology witnesses; the 17 built operations cover floors, roofs, columns, doors, rooms, wall openings, stairs, direct shapes, curtain grid lines and panels, family placement, fixture modification, conduits, placeholders and flexible ducts and pipes. Earlier live runs on 2026-07-26 and 27 established arc walls, slanted columns and roof slopes and repaired the reverse path's collector, which had returned nothing on every version until then. Numbers that appear in the project's README without an artefact — a distribution of 85,374 production compile errors, a count of 1,056 compilations, and “31 of 31 witnessed operations”, which the contemporaneous record gives as 30 of 31 — are not used. The public catalogue `paper/exp_v6_live/live_execution_catalogue.md` states which records exist: the two operation matrices and the compile gate are native records that are not shipped, whereas the count of operations proven live before 2026-07-27, the bureau runs and the 92.83% re-lift rest on project notes. Table S8 is the table view of main-paper Figure 7a.

**Table S8.** The evidence ladder on the snapshot.

| Rung | Artefact | Reproduced here | Level |
|---|---|---|---|
| Emitted | C# text; golden programs | yes | IMPL |
| Compiled | compile service vs Revit API assemblies 2021–2026 | no; 1,242/1,242 on 2026-07-27 | ROSLYN |
| Dispatched | injected callback; HTTP client; C# connector with operation journal and document barrier | contracts and fixtures | IMPL; project-reported |
| Committed | commit-status check; typed receipt; connector receipts | no; 18/20 + 8 (2023); 17/21 (2026) | LIVE |
| Observed | in-transaction witness; census probe; re-extraction | fixtures; 92.83% re-lift of a 90,758-element capture | TEST-OFF; LIVE |
| Intent verified | `PROGRAM` vs `PARSE` verdict; end-to-end walkthroughs | comparator only; walkthroughs 8/8 and 11/11 project-reported | IMPL; LIVE |

## S4 The coordination layer in detail

**Building graph.** `graph_from_l0` builds a typed graph from an extracted document with one node type addressed by the document's own element id, so rooms, levels, grids and elements share one address space. Edges carry one of eleven relations — hosted in, hosted in a linked document, placed on a datum, on level, level above, bounds room, joined to, joined at end, top constrained to level, bounded by the same wall, opening point touches room — and a modality on an orthogonal axis (proven, possible, refuted, unresolved target). A host absent from the snapshot does not vanish: the edge is kept with modality unresolved and a typed reason. On the project's synthetic two-level document (not the one in main-paper Figure 8) the graph has 11 nodes from 11 rows and 24 edges over eight relation kinds, with the census balanced, no refusal and three of the four modalities exhibited on one relation (36 ms); the project reports 76 of 76 corpus documents built with no refusal (1,576,343 nodes; 1,843,930 edges).

**Clash detection with an exact phase.** The pipeline builds conservative hulls, enumerates candidate pairs on a spatial grid and reports a hull relation (overlap, contact, separated) and a verdict (confirmed, possible); a hull that cannot be built enters the census as a refusal with a kind. The exact narrow phase computes body intersections in the OCCT kernel with a closed set of refusals. On the project's six-element scene (again not the one in Figure 8), five are hulled and one refused with a kind; the default mechanical-versus-structural filter reports two findings (a duct against each wall, signed distances −200 and −100 mm, verdict possible because inner evidence is absent), all pairs report three, and the separated walls report none. Exact clash, repair and refinement deviation are kernel-gated on the evaluation machine; 57 hull-level clash tests and 5 exact-distance witness tests pass.

**The geometry loop without a host.** Bodies authored in the sandbox participate in clash analysis with the same body, frame and tolerance used for emission; a finding is repaired in the authoring project by a proposal, and before anything is written the repair is re-analysed on the candidate merge and refused by name if it opens a new conflict, worsens an existing one or fails to separate the target pair. The project's end-to-end instrument runs eight steps of this loop — create, open, geometry change, refinement, analysis and repair, team edit, forced termination, continuation — and reports 8 of 8; a control residential complex with a curved fifteen-degree loft, terraces and a third reopening reports 11 of 11. Both are project-reported; on the evaluation machine the instrument stops at its first geometry step and names `kernel_unavailable` rather than passing.

**Refinement and lineage.** A detailed section remembers its concept; the residue that the translation could not express and the deviation of derived bodies from the source are numbers with addresses, with refusal codes `KIR-R004` to `R015` for cases the kernel cannot settle. A later change to the source partitions every derived output into recomputed, preserved and needs-decision — three sets whose union must be the whole instance — and raises a question with choices where geometry cannot decide; answering it is itself a proposal. A change to an atrium radius reaches the twenty-one walls detailed from it.

**Offline editing of captured buildings.** A building captured from the host is opened without it, described field by field in one of four states (represented, approximate, source data, unknown), edited by element address and saved atomically with a per-field ledger of losses; the export compiles. On the project's benchmark capture of 4,223 elements and 58,451 non-empty fields, the ledger reports 26,282 represented, 1,647 approximate, 2,940 source-data and 27,582 unknown fields; 26 tests pass.

**Native updates and the document barrier.** Updates the host cannot perform inside the program-wide transaction (a sketch edit requires no active transaction) are emitted as self-owning transaction units with observation before and after, and a type change that the host may implement by creating a new element is read back through an explicit identity-replacement row. On the host side, the compiler host binds generated code to exactly one document by Roslyn semantic analysis: any expression producing a `Document` other than the bound parameter is refused, and a change that arrives in another document's undo stack is a terminal refusal.

**Stages, disciplines, the judge and the agent's environment.** A stage profile suspends rules whose prerequisites a concept model does not carry (14, 11 and 11 of 20 rules are judged under the as-built, static design-intent and dynamic profiles). Every operation belongs to one of six disciplines, derived from the categories it creates through a table cross-checked between two independent carriers. The minimum-ceiling-height rule on two subjects reports two violations at 2,500 mm and none at 2,286 mm. An agent authors in a sandboxed runtime whose namespace is the authoring course's language module — sixteen lessons of 39,905 characters, each under the sandbox's 3,300-character output budget — and the MCP door exposes seven tools, five of them without a host, every reply stating whether anything was written.

**The bureau.** A mission is a plan of turns, each naming its worker, instance, declared outputs and backward-only dependencies; `advance` executes exactly one next turn. Figure S1 shows the claim, proposal and conflict protocol. A long task continues from its memory — the journal of accepted turns and recorded decisions, not a chat transcript. Budget is a team ceiling in calls and tokens; the stop check precedes the budget check, so a stopped team spends nothing. Responders are external: a file-exchange provider hands requests to language-model agents through a directory and validates their answers (schema, non-empty text, usage within budget), and a tape provider replays recorded exchanges for tests. The project's notes report three runs with two external responders over the file exchange on 2026-09-07 and 08, each passing 9 of 9 checks; their raw records were not located for the public bundle.

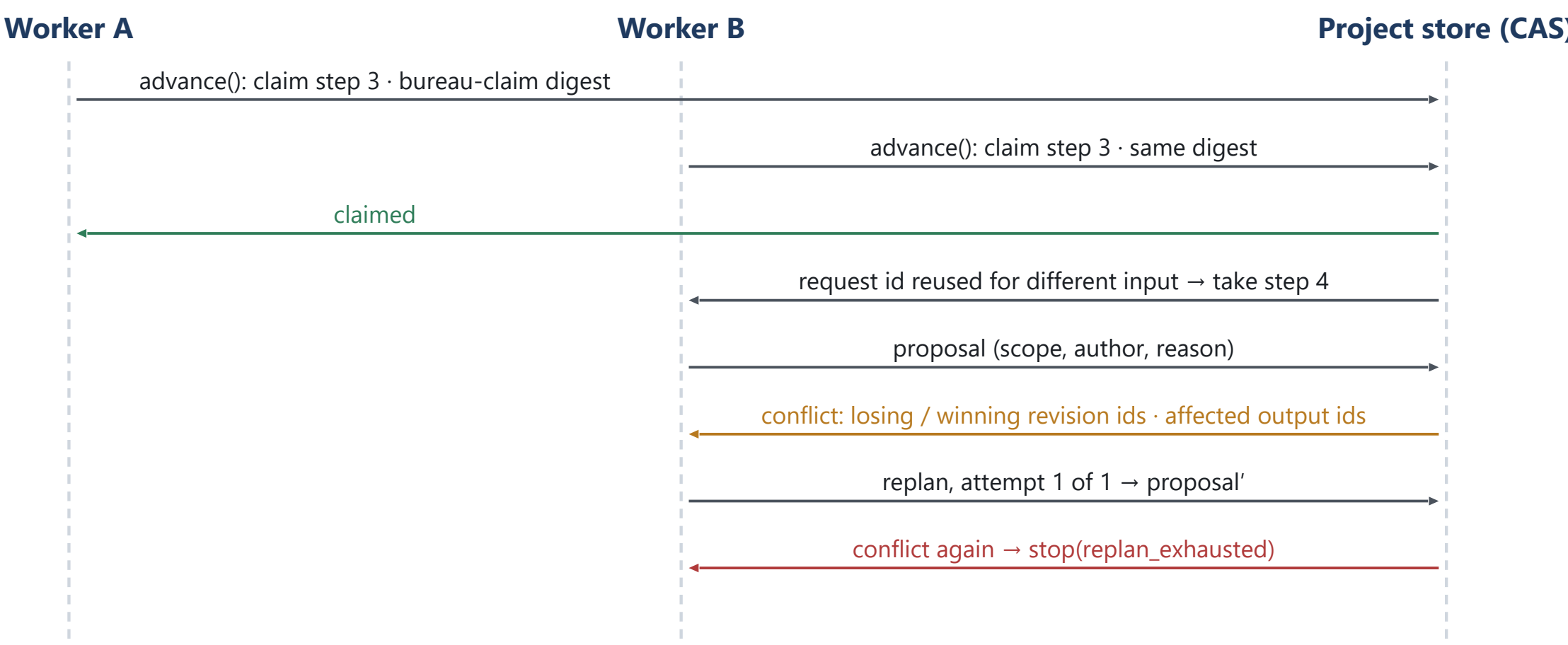

**Figure S1.** Two workers of the bureau share one plan through compare-and-swap claims; a conflict is returned as a replan request with addresses, and a second conflict stops the team by name.

**Directions.** The foundation admits extensions with a measurable outcome in the vocabulary of the paper (refused, recorded, unread, conflicting, unwitnessed): an IFC writer for the expressible subset, with the contracts that have no IFC analogue reported as named absences, and a declarative backend description that retires the inline version branches; scan-to-BIM front ends emitting programs under the census invariant [S1]; quantity take-off attributed to authored, defaulted and compiler-derived values [S2]; construction sequencing from declared effects and recovered host relationships [S3]; compliance checking with coverage statements [S4, S5]; clash avoidance across disciplines before execution [S6]; and the RQ2 negative corpus as a benchmark for generators.

## S5 RQ7 protocol (not executed)

**Producers.** The same language model with the same prompt budget, once instructed to write Revit API C#, once to write Python against `kir.dsl`; both receive the same snapshot facts as text. **Tasks.** Ten authoring tasks of increasing statefulness: (1) a room; (2) a room with openings; (3) a floor with holes; (4) a two-storey stack; (5) a family placement that needs a loaded symbol; (6) an edit of an existing wall; (7) an edit of a hosted door after its wall moved; (8) a deletion with dependants; (9) a change of level elevation; (10) a conflicting concurrent edit. **Scoring.** Per task, by predicates fixed before the run: (a) compiled by Roslyn for the target version; (b) executed and committed in a live Revit of that version; (c) post-conditions read back match the stated intent within tolerance; (d) for edit tasks, untouched elements keep their element ids; (e) number of repair rounds; (f) whether the arm can show effects before execution; (g) state of the document after a forced unknown outcome, with the bridge killed mid-transaction. A compile-time refusal in the KIR arm counts as a failure unless repaired within the same budget. Source and token size are recorded, not scored. At least two Revit versions. **Why it is decisive.** The existing evaluation establishes that the mechanisms exist and handle failure; only this comparison tests the paper's central comparative hypothesis on the second and third edit, preservation, repair and recovery after an uncertain transaction. **Status on the snapshot:** specified, not run; no comparative number is reported.

## S6 Reproduction and cross-environment re-derivation

**Environment.** The KIR 0.8.1 source tree; Python 3.12.13 with numpy and httpx; `PYTHONPATH=<tree>` and `PYTHONNOUSERSITE=1`, with `python -c "import kir; print(kir.__file__)"` checked to print the tree path before any number is trusted. The OCCT kernel is required for the exact clash, repair and deviation tests, which otherwise skip by name. No Revit, no bridge, no network. The same scripts were first run on 0.6.0. **Scripts, per research question** (in the repository's `paper/` directory; paths are relative to it).

**RQ1** — `scripts/rq1_abcdf.py` (name sets, signatures, schema, documentation); `scripts/rq1_e_matrix.py` (83 × 6 compilation attempts); `scripts/rq1_g_string_literals.py` (slot-reference scan). Outputs: `rq1_*.json`.

**RQ2** — `build_corpus.py` writes the corpus and manifest; `run_corpus.py` calls `compile_program(program, revit_version="2026", snapshot=SNAP, bulk=True)` per case and records ok, codes, first message line, operation and field named, C# length, exception and wall time; `dsl_probes.py` writes the DSL probes.

**RQ3** — `scripts/rq3_h_i_l.py` (five programs — the four-wall README room, the room with a door and the tower's three programs — × six versions; three repeated emissions with SHA-256); `scripts/rq3_i_diffs.py` (diffs between consecutive versions).

**RQ4** — `exp_v6_persist/experiment.py` (project store); `recon_persist/experiment.py` (journal of observed state), with the listed test files.

**RQ5g, RQ6** — coordination scripts and outputs in `exp_coord/` (graph, clash, stages, course); the live-execution catalogue in `exp_v6_live/live_execution_catalogue.md`. The bureau, refinement, capture and exact-clash invocations were not located and are not included.

**Cross-environment re-derivation.** For this revision the key counts were re-derived from a second source archive of the same release (the GitHub archive of KIR 0.8.1 of 2026-09-08) on Windows with Python 3.14, by direct calls into the package (Table S9). This checks the reported numbers against a second source tree and environment; it is not a third-party replication. All structural counts agree exactly with the snapshot, and the RQ3 re-run confirms its denominator: five programs × six versions, 30 of 30 emissions without diagnostics. The tower's character counts differ by a few characters between the two trees, and the ratio is unchanged. The worked example of main-paper Figure 2 and the tower geometry of Figure 1 were produced in the same way, and main-paper Figure 4 is generated by the enumeration in the table's fourth row.

**Table S9.** Cross-environment re-derivation from a second source archive of KIR 0.8.1 (Windows, Python 3.14). "Snapshot"

is the value reported in the evaluation.

| Quantity | How it was re-derived | Re-derived | Snapshot |
|---|---|---|---|
| operations by family; writing operations | iterate `spec.OPS` | 83 = 71 + 6 + 6; 77 | 83 = 71 + 6 + 6; 77 |
| effect kinds | `OpSpec.effect` over `spec.OPS` | create 71, read 6, mutate 5, delete 1 | 71 / 6 / 5 / 1 |
| parameter kinds; reference kinds | `PARAM_KINDS`; `ReferenceKind` | 39; 8 | 39; 8 |
| outcome space; retry classes; first-violated invariant | construct all 100 `ProgramOutcome` triples | 31; 14 / 13 / 4; 20 / 5 / 0 / 16 / 11 / 14 / 3 | 31; 14 / 13 / 4; 69 refused |
| witness obligations over writing operations | obligation table in `translation_cert.py` | 377 (2 to 13 per operation) | 377 |
| named absences: exempt / absence / caveat | clause notes in `translation_cert.py` | 38 over 26 operations: 27 / 8 / 3 | 38 over 26: 27 / 8 / 3 |
| tower: programs; authored and expanded operations; elements | run `examples/tower_numpy.py` | 3; 6 → 840; 780 | 3; 6 → 420 (see note) |
| tower: KIR vs emitted C# (Revit 2023), characters | same script | 11,264 vs 3,709,229 (329×) | 11,263 vs 3,709,235 (329×) |
| RQ3: programs × versions; emissions without diagnostics | README room, room with door, the tower's three programs; Revit 2021–2026 | 5 × 6; 30 / 30 | 30 / 30 |
| RQ3: room, KIR vs emitted C# (Revit 2023), characters | same run | 612 vs 34,420 | 612 vs 34,420 |

The source manuscript described the tower's three programs as “expanding to 420 operations”; both the project's README (dated 2026-09-04, before the snapshot) and the re-run give 840 expanded operations (60 levels, 60 slabs and 720 columns) and 780 elements, and the main paper uses these values.

## S7 Development method and internal audits

The project applies the principle of main-paper Section 2 to its own claims through three instruments. *Measured, not promised:* the README's table of numbers is read by a shipped instrument that recomputes each recomputable row and exits non-zero on divergence; on the snapshot 19 rows matched, 3 diverged by one (a module, a test file and a parsed file added after the README was regenerated) and 4 are labelled historical. *The stranger gate:* the owner's measure of progress is whether a person with no host, no ports and no Revit can install the package into an empty environment and obtain correct answers from twenty-four representative commands, each classified as truthful, a false zero or a crash; it read 18 / 3 / 3 on 2026-08-30. *Agreements:* a registry of self-consistency checks between code paths that must agree, each with the incident that motivated it and a mandatory demonstration that it can fail.

Between 2026-08-29 and 2026-09-04 the project ran an audit of its own correctness by a fleet of language-model agents — a lead owning the findings ledger, a measurer, and seven workers each owning a code area. The ledger holds 548 findings, 188 at the highest priority (135 fixed and 42 open at 0.6.0); each fix carries an executed discriminator, and an additive gate checked that no public name vanished across 48 commits and 97 patches. A recheck of the 127-item correctness subset confirmed 97, confirmed 25 with a caveat and refuted 5. A second audit on 2026-09-06 produced twelve reports whose findings (among them a connector policy admitting a transaction on a foreign document and a write without an operation id producing a double create) were closed before the snapshot. Both audits are project-internal evidence of defect classes, not independent validation.

## S8 Further limitations

Beyond main-paper Table 3: diagnostics exist only in Russian (codes and field names are language-neutral), as do the room classifier and the authoring course, so a bureau in another jurisdiction needs externalised norms and an English authoring surface. Version branching is inline per emitter. The certificate is disabled by default in the serving path. The observed-state journal is identity-free, so authored identity is kept only in the store. The building graph and clash hulls are built from extracted state or authored bodies, not yet from a grounded program, so an agent cannot learn what a proposed wall bounds before it is materialised. Constraints between disciplines are conventions of declared scope rather than checked contracts. Two fidelity proofs of the reverse path are weaker than designed (project findings F-306 and F-281).

Two representations of the unconfirmed state coexist — the outcome space and the boolean fields of the receipt, which the recovery subsystem uses. The effects scheduler is implemented and tested but not wired. The written specification promises reference kinds, per-operation fields and a grammar for constrained decoding that the code does not implement; the paper describes the code.

## References cited in the supplement

[S1] P. Tang, D. Huber, B. Akinci, R. Lipman, and A. Lytle. Automatic reconstruction of as-built building information models from laser-scanned point clouds: a review of related techniques. *Automation in Construction*, 19(7):829–843, 2010.

[S2] M. Valinejadshoubi, O. Moselhi, I. Iordanova, F. Valdivieso, and A. Bagchi. Automated system for high-accuracy quantity takeoff using BIM. *Automation in Construction*, 157:105155, 2024.

[S3] O. Doukari, B. Seck, and D. Greenwood. The creation of construction schedules in 4D BIM: a comparison of conventional and automated approaches. *Buildings*, 12(8):1145, 2022.

[S4] C. M. Eastman, J. Lee, Y. Jeong, and J. Lee. Automatic rule-based checking of building designs. *Automation in Construction*, 18:1011–1033, 2009.

[S5] Z. Zhang, L. Ma, and T. Broyd. Rule capture of automated compliance checking of building requirements: a review. *Proceedings of the Institution of Civil Engineers — Smart Infrastructure and Construction*, 176(4):224–238, 2023.

[S6] A. O. Akponeware and Z. A. Adamu. Clash detection or clash avoidance? An investigation into coordination problems in 3D BIM. *Buildings*, 7(3):75, 2017.